\documentclass[pmlr]{jmlr}

\RequirePackage{graphicx}
 \usepackage{booktabs}
 \usepackage{array}
 \usepackage{wrapfig}
\usepackage{longtable}
\usepackage{booktabs}
\usepackage{siunitx}
\usepackage{graphicx}
\makeatletter
\renewcommand*{\@subfigurelabel}[3]{#1\subfigurelabel{#2}}
\newcommand*{\arxivsubref}[1]{{%
  \def\@subfigurelabel##1##2##3{\subfigurelabel{##2}}%
  \ref{#1}%
}}
\makeatother
\newcommand{\subref}[1]{\arxivsubref{#1}}
\usepackage{makecell}
\usepackage{enumitem}

\makeatletter
\def\set@curr@file#1{\def\@curr@file{#1}} 
\makeatother
\usepackage[load-configurations=version-1]{siunitx} 

\theorembodyfont{\upshape}
\theoremheaderfont{\scshape}
\theorempostheader{:}
\theoremsep{\newline}

\jmlrpages{} 
\jmlrproceedings{PMLR}{Proceedings of Machine Learning Research}
\jmlrvolume{340}
\jmlryear{2026}
\jmlrworkshop{Machine Learning for Healthcare}

\title[PAFIR: Personalized and Adaptive 
Fall Risk Identification and Prevention]{
Adaptive Multi-Agent Feature Selection for Personalized Fall Risk Prevention
}

\author{\Name{Chang Liu}
       \Email{chang.liu@ucf.edu}\\ 
       \addr School of Data, Mathematical, and Statistical Sciences\\
       University of Central Florida\\
       Orlando, FL, USA 
       \AND
       \Name{Ladda Thiamwong}
       \Email{Ladda.Thiamwong@ucf.edu}\\ 
       \addr College of Nursing\\
       University of Central Florida\\
       Orlando, FL, USA 
       \AND
       \Name{Yanjie Fu}
       \Email{Yanjie.Fu@asu.edu}\\ 
       \addr School of Computing and Augmented Intelligence\\
       Arizona State University\\
       Tempe, AZ, USA 
       \AND
       \Name{Rui Xie\textsuperscript{\(\dagger\)}}
       \Email{Rui.Xie@ucf.edu}\\ 
       \addr School of Data, Mathematical, and Statistical Sciences,  College of Nursing \\
       University of Central Florida\\
       Orlando, FL, USA 
       } 

\begin{document}

\maketitle

\begingroup
\renewcommand{\thefootnote}{\(\dagger\)}
\footnotetext{Corresponding author.}
\endgroup

\begin{abstract}
Falls among older adults represent a major public health challenge driven by complex, time-varying interactions across multiple risk domains. Effective fall risk factor identification requires learning from heterogeneous longitudinal data while accounting for sparse and delayed fall-related outcome events. However, existing approaches are largely static and fail to adaptively model evolving, individualized risk factors across modalities and time.
We propose \textbf{PAFIR}, a \textbf{P}ersonalized and \textbf{A}daptive \textbf{F}eature selection framework for fall risk \textbf{I}dentification and p\textbf{R}evention, which formulates adaptive feature selection as a reinforcement learning problem over longitudinal multimodal health data. PAFIR jointly models structural dependencies among correlated assessment variables and temporal dynamics in wearable-derived physical activity data, and learns adaptive selection policies across repeated study visits using reward signals derived from sparse fall incidence outcomes.
We apply PAFIR to data from the {P}hysio f{E}edback {E}xercise p{R}ogram (PEER) cluster-randomized trial. Experimental results demonstrate that PAFIR more effectively captures longitudinal and structural patterns of feature relevance than state-of-the-art baselines, and enables dynamic, subject-specific feature selection. By adapting selected features over time, PAFIR supports more timely and personalized fall prevention strategies.
\end{abstract}

\section{Introduction}
\label{sec:intro}

Fall prevention among older adults remains an urgent and complex public health challenge. In the United States, falls are the leading cause of fatal and non-fatal injuries among adults aged 65 and older, resulting in over $38{,}000$ deaths annually and substantial healthcare costs~\citep{cdc_falls_2025, florence2018medical}. Beyond acute injury, falls often trigger long-term functional decline, loss of independence, and diminished quality of life~\citep{paliwal2017chronic}. As populations age and care resources become increasingly constrained, there is a critical need for timely, scalable, and personalized strategies to identify fall risk and intervene before adverse events occur. 

Recent advances in mobile and smart health technologies enable continuous monitoring of gait, activity, balance, and contextual behaviors, creating new opportunities for early fall risk detection~\citep{pfortmueller2014reducing, giovannini2022falls, mortazavi2023low}. However, translating these heterogeneous, high-dimensional data sources, which include continuous sensor streams, multi-time-scale signals, and structured survey-based measurements, into effective prevention remains challenging. 

 \begin{figure}[htbp]
   \centering 
   \includegraphics[width=\linewidth]{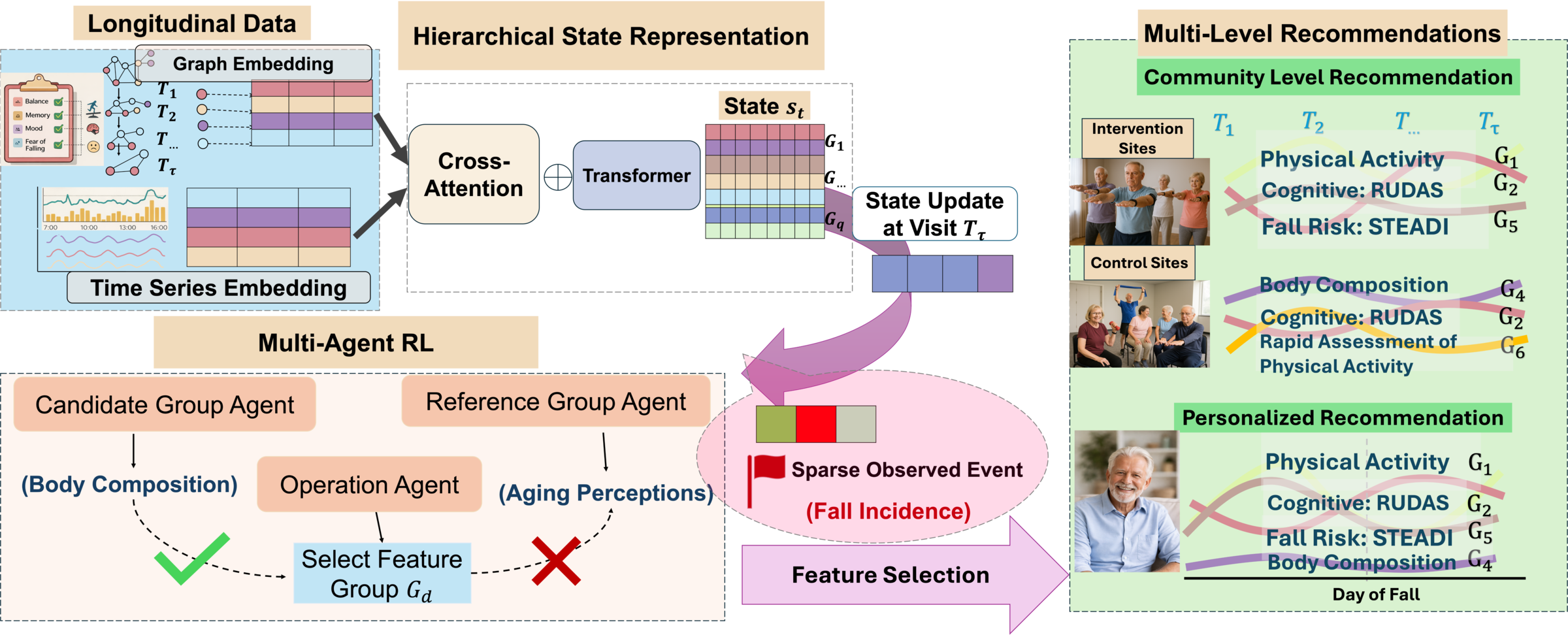} 
   \caption{\small Overview of the \textbf{PAFIR} framework. Heterogeneous \textbf{longitudinal data} from multiple modalities are integrated to construct a \textbf{hierarchical state representation} of the individual state. This state is iteratively updated across clinical visits and in response to sparsely observed events, i.e., fall incidence. A \textbf{multi-agent reinforcement learning} module then operates on the evolving state to perform adaptive \textbf{feature selection} at both group and individual levels. The resulting learned policies support \textbf{personalized and community-level recommendations} for fall risk identification and prevention.}
   \label{fig:overview} 
 \end{figure} 

A critical insight, however, is that fall risk is inherently dynamic and highly individualized, shaped by both objective physical capacity and subjective risk perception. Older adults may underestimate their physiological vulnerability or, conversely, restrict activity due to fear of falling despite preserved function~\citep{duncan1993physiological, delbaere2010determinants, cuevas2017balance, liu2025effectiveness}. Such misalignment between the \textit{Body} (mobility, balance) and the \textit{Mind} (confidence, cognition, and risk perception) can lead to maladaptive behaviors, delayed risk recognition, and reduced engagement with preventive care~\citep{he2022human, wang2024age, maruszewska2025risk}. These complexities make effective prevention not only a problem of detecting risk signals, but also of adaptively determining \emph{which} features are most informative and \emph{when} to intervene for each individual~\citep{dautzenberg2021interventions, pillay2024falls}. 

\begin{wrapfigure}{r}{0.5\linewidth}
    \vspace{-0.15in}
    \centering
    \includegraphics[width=\linewidth]{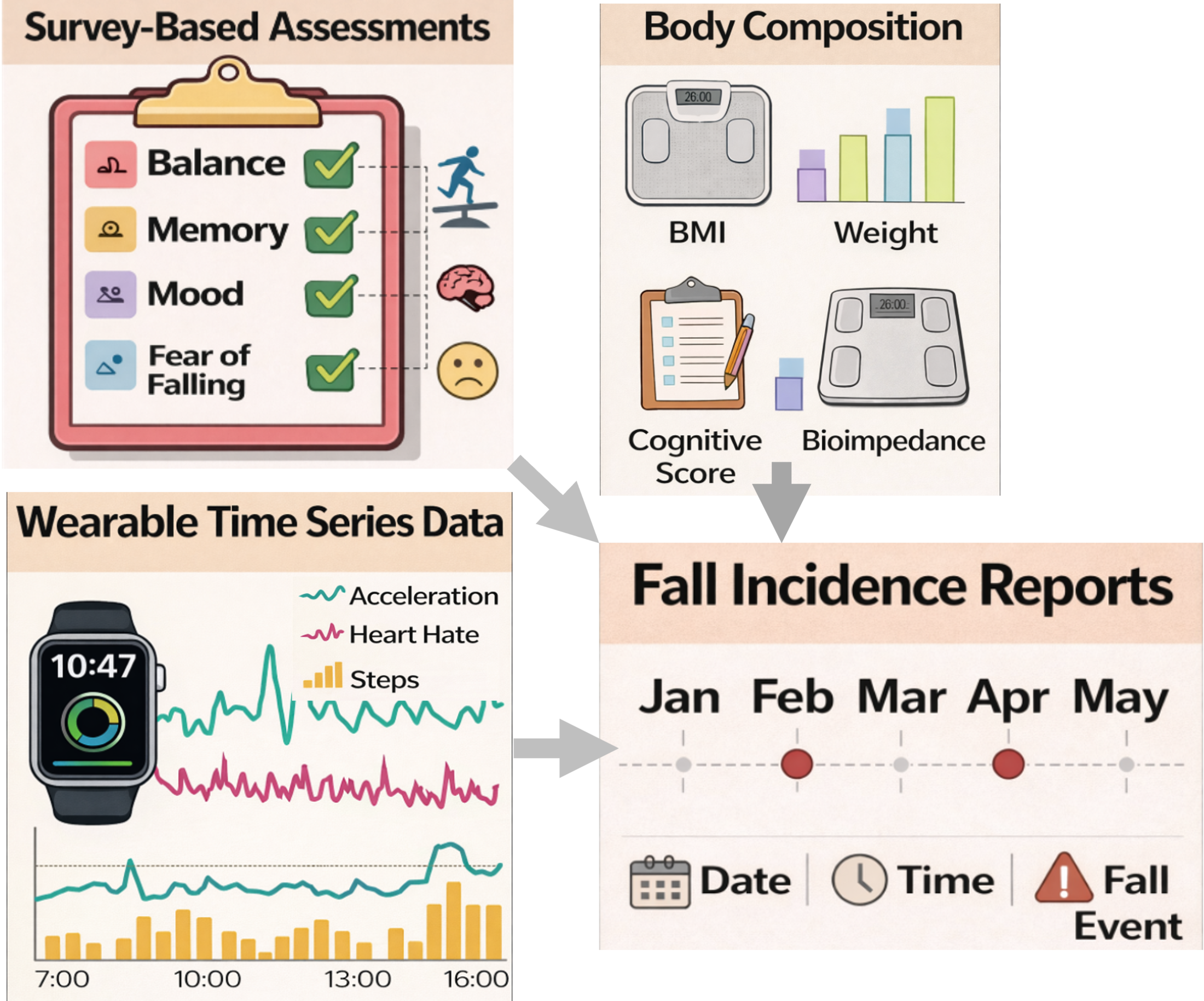}
    \caption{\small 
    Illustration of the multimodal longitudinal PEER dataset across visits (T1-T4), including (1) survey-based fall risk assessments, (2) high-frequency wearable time-series data capturing physical activity, and (3) instrument-based measurements such as body composition. \textbf{Fall incidence}, the primary outcome, is recorded as sparse, time-stamped events.
    }
    \label{fig:data_structure}
    \vspace{-0.2in}
\end{wrapfigure}

Although many individual fall risk factors have been studied, how physiological, psychological, and behavioral factors interact and co-evolve over time remains poorly understood. Most existing approaches model these factors in isolation or assume static relationships across visits~\citep{gong2024hn, yassine2021intelligent, theng2024feature}, limiting their ability to capture dynamic longitudinal interactions between physical capacity, risk perception, and behavior. The {P}hysio f{E}edback {E}xercise p{R}ogram (PEER) study~\citep{Thiamwong2023} is a technology-based intervention, specifically, designed to address this mismatch, integrating real-time physio-feedback, cognitive reframing, and peer-led exercise to jointly engage physical and psychological fall risk domains. 
This study offers rich, longitudinal multimodal data spanning structured survey-based assessments, clinical measurements, fall incidence, and wearable sensor-derived time-series data, as visualized in Fig.~\ref{fig:data_structure}, providing a comprehensive data foundation for data-driven fall risk identification and prevention. This gap motivates data-driven methods that integrate multimodal signals and adaptively model evolving risk factor interactions, framing personalized fall risk identification as an adaptive feature selection problem over heterogeneous physiological, behavioral, and contextual data, as illustrated in the overview of the proposed PAFIR framework in Fig.~\ref{fig:overview}.

Emerging developments in reinforcement learning (RL) for healthcare have shown strong promise for adaptive feature selection as a sequential decision process, enabling models to update feature relevance based on observations and clinical feedback~\citep{yu2021reinforcement, komorowski2018artificial}. Despite these advances, existing feature selection systems continue to face fundamental limitations when deployed in real-world healthcare settings, particularly for fall prevention.

A primary challenge in fall risk factor identification arises from the integration of highly diverse data sources, including self-reported surveys~\citep{ritchey2022steadi}, body composition measurements~\citep{nguyen2024unveiling}, muscle strength and balance evaluations~\citep{mcmanus2022development}, daily physical activity captured via wearable devices~\citep{howcroft2017prospective, liu2025diffusion}, and gait and posture assessments~\citep{lim2024fall}. Most existing studies focus on only a subset of these data types or rely on simplified fusion strategies~\citep{gonzalez2024applications}, limiting their ability to capture cross-modal interactions and evolving risk patterns. 

Additionally, fall risk modeling is further complicated by the dynamic and heterogeneous nature of activity and risk signals in older adults. Wearable sensors produce continuous, minute-level physical activity data~\citep{alsadoon2024architectural}, while fall outcomes are sparse, delayed, and irregular~\citep{tonchoy2024mental, li2025falls}, complicating temporal alignment and longitudinal analysis. Gradual changes in mobility, cognition, and psychological state may develop over extended periods, whereas acute events can induce abrupt shifts in risk. However, learning from sparse outcome signals for fall incidence remains a major challenge. For example, in the PEER study~\citep{Thiamwong2023}, fall events are inherently rare, with only 69 events observed across 1364 visit-level observations and many participants experiencing no falls throughout the entire trial. When the reward is defined solely based on the binary occurrence of a fall, the RL agent receives informative feedback in only a small fraction of training steps~\citep{wang2020reinforcement, ecoffet2021first}, corresponding to a setting with limited and delayed feedback where informative signals are difficult to propagate across time~\citep{wang2023addressing}. 

Fall-risk research has employed a broad range of statistical and machine-learning methods for identifying relevant risk factors. Marginal filtering and importance-ranking approaches typically evaluate variables separately, whereas multivariable methods, including LASSO~\citep{tibshirani1996regression} and Group LASSO~\citep{yuan2006model}, perform joint selection under sparsity constraints. Established longitudinal and survival-analysis approaches can also accommodate repeated measurements and event outcomes. For example, joint models~\citep{wang2025joint,elashoff2016joint,crowther2016joint} link longitudinal biomarker trajectories with time-to-event outcomes, while frailty models~\citep{hougaard1995frailty} account for unobserved heterogeneity. These methods are particularly appropriate when the primary objective is to estimate time to fall, covariate effects, or the probability of falling within a prespecified prediction horizon~\citep{pepe2003statistical,suresh2022survival}.

To address the challenges above mentioned, we introduce \textbf{PAFIR}, a \textbf{P}ersonalized and \textbf{A}daptive \textbf{F}eature Selection Fall Risk \textbf{I}dentification and p\textbf{R}evention framework (Fig.~\ref{fig:overview}) that formulates fall risk factor identification as a multi-agent reinforcement learning feature selection problem over longitudinal multimodal health data. 
PAFIR models heterogeneous clinical assessments and high-resolution activity data while accounting for temporal dynamics and individual variability. By jointly capturing structural relationships among risk factors and their evolution over time, PAFIR learns adaptive feature relevance from longitudinal data. 
An RL policy guided by sparse fall outcomes and proxy fall risk appraisal measures enables personalized and group-aware identification of evolving fall risk factors to support timely prevention.

\paragraph{Contributions Overview}
\begin{itemize}[noitemsep,topsep=1pt,leftmargin=*]
    \item \textbf{Hierarchical structure for fall risk feature selection in the clinical trial}: 
    We embedded a hierarchical structure in a topology-aware reinforcement learning framework to effectively capture and represent multi-level dependencies among features, which is crucial for fall risk screening, where factors are naturally organized hierarchically. 
     For example, within fear-of-falling assessments, questionnaire items measuring activity avoidance and balance confidence belong to the same higher-level domain while reflecting distinct behavioral and perceptual contributors to fall risk.
    \item \textbf{Adaptive comparison-driven feature selection mechanism}: We propose an adaptive comparison-driven feature selection mechanism that decomposes the selection process into candidate generation, reference-based comparison, and iterative refinement. Instead of evaluating features independently, the method assesses their incremental contribution by comparing candidate feature groups against reference groups, enabling more reliable fall risk factor identification under complex healthcare settings. 
  \item \textbf{Sparse-aware rewards design with clinical proxies}: We introduce a sparse-aware reward design that augments sparse fall-event signals with clinically informed proxy outcomes (i.e., FES-I and BBS), allowing the PAFIR framework to capture broader fall-risk patterns and improve the robustness of feature selection. 
\end{itemize}

\paragraph{Generalizable Insights about Machine Learning in the Context of Healthcare}
The proposed framework can generalize to settings that require identifying clinically relevant risk factors from heterogeneous and multimodal data, including structured clinical assessments, behavioral measures, and high-frequency sensor signals. Such settings are common in healthcare, where patient information is collected across diverse modalities and scales, and effective modeling requires integrating these sources while preserving their structural relationships.
Moreover, the approach accommodates both discrete and continuous outcomes, ranging from binary events (e.g., disease onset or adverse events) to continuous measures (e.g., functional decline or physiological indicators), enabling a unified and flexible framework for modeling feature relevance across diverse clinical settings. In addition, the framework naturally extends to sequential decision-making settings, where feature relevance evolves over time and depends on prior observations and selections. This is particularly important in longitudinal healthcare data, where risk factors may change gradually or abruptly, and adaptive modeling is required to capture temporally evolving patterns. Finally, the use of informative feedback mechanisms, such as incorporating proxy or auxiliary signals, enables effective learning in scenarios with sparse or delayed outcomes. Such conditions are prevalent in healthcare applications, where clinically meaningful events are often rare, and leveraging additional signals can improve stability and guide learning toward meaningful trajectories. Together, these insights suggest that integrating multimodal data, supporting diverse outcome types, and enabling adaptive sequential learning are key to developing generalizable machine learning methods for healthcare.

\section{Related Work}
Fall-risk assessment traditionally relies on clinically interpretable screening and functional measures, including the Timed Up and Go test, balance assessments, and STEADI-related indicators. With the increasing availability of wearable sensors and longitudinal cohorts~\citep{picerno2021wearable}, machine-learning studies have incorporated physical activity, gait, cognitive, psychological, and functional features for broader risk characterization~\citep{fan2024predicting}. However, many existing applications still use fixed feature sets or collapse longitudinal observations into visit-level summaries~\citep{chantanachai2021risk,rykov2021digital}.

Statistical methods provide an established framework for longitudinal fall-outcome analysis. Time-varying and recurrent-event survival models accommodate changing covariates and repeated falls, frailty models account for unobserved heterogeneity~\citep{hougaard1995frailty}, joint models link longitudinal biomarker trajectories with event outcomes~\citep{wang2025joint,elashoff2016joint,crowther2016joint}, and competing-risk models address terminal events such as death~\citep{li2022comparison}. Feature selection can be incorporated through LASSO, Group LASSO, longitudinal penalization, or boosting~\citep{tibshirani1996regression,yuan2006model,xu2015longitudinal,dong2022neural}. These approaches are well suited to estimating covariate effects, event risk, or survival probabilities under a specified outcome model~\citep{kalbfleisch2023fifty,suresh2022survival,smith2022scoping}. PAFIR instead focuses on adaptively updating feature sets from visit-level assessments and wearable sequences as new fall or proxy feedback becomes available.

Healthcare data are inherently multimodal, irregular, and hierarchically structured, combining sparse events, wearable sensor streams, and structured assessments~\citep{ghat2023future, franklin2024modernizing}. Many clinical surveys exhibit hierarchical organization, such as cognitive domains composed of multiple subscales~\citep{rosellini2021developing}, yet most models ignore these dependencies, leading to information loss~\citep{mukherjee2023encoding}, particularly in longitudinal settings. Recent work on hierarchical structural encoding~\citep{luo2024enhancing} motivates the need for personalized, temporally adaptive modeling, which our framework aims to address.

To address these limitations, recent work has explored dynamic and adaptive feature selection strategies. Reinforcement learning has emerged as a promising approach for dynamic feature selection and sequential decision-making in healthcare, with applications such as adaptive screening~\citep{yu2021reinforcement} and treatment optimization in critical care~\citep{komorowski2018artificial}. For feature selection, TTG~\citep{khurana2018feature} formulates feature transformation as a graph construction problem and applies RL to search for optimal subsets, but lacks temporal modeling and assumes static relationships. Structure-Aware Transformer (SAT)~\citep{chen2022structure} improves graph-based modeling by extracting node-centric subgraphs before attention computation, yet it does not capture evolving longitudinal or personalized dynamics. Topology-Aware Reinforcement Learning (TAR)~\citep{ying2024topology} introduces dynamic graph adaptation through reinforcement-based feature space reconstruction. 

However, existing adaptive feature-selection methods provide important foundations for graph-based or sequential feature-space optimization, but they are not designed specifically for the combination considered here: clinically defined hierarchical feature groups, visit-level structured assessments, minute-level wearable sequences, asynchronous information updates, and sparse fall outcomes supplemented by proxy feedback. PAFIR integrates these elements within a unified adaptive feature-selection framework for longitudinal multimodal health data.

\section{Clinical Trial for Fall Prevention: \textbf{P}hysio-f\textbf{E}edback \textbf{E}xercise p\textbf{R}ogram}
We are motivated by the heterogeneous and longitudinal data structure (Fig.\ref{fig:data_structure}) from the \textbf{P}hysio-f\textbf{E}edback \textbf{E}xercise p\textbf{R}ogram (PEER) study~\citep{Thiamwong2023}, a two-arm clustered randomized controlled clinical trial designed to evaluate a technology-assisted, body and mind intervention for fall prevention and physical activity promotion among older adults in a free-living environment (clinicaltrials.gov ID: NCT05778604). The analyzed cohort includes 341 participants, aged 61 to 89 years, from 15 senior living sites in Central Florida. 
In this two-arm cluster randomized trial, senior living centers were randomly assigned to either the PEER intervention arm or the control arm, with all participants within each center receiving the same allocation.

Each participant completed four clinical assessments: baseline at week 0 (T1), post-intervention at week 9 (T2), 2 months (60 days) post-intervention (T3), and 6 months post-intervention (T4). All assessments and wearable-monitoring periods followed the scheduled visit protocol and were not triggered by fall events. Following each scheduled visit, participants were asked to wear the accelerometer for seven consecutive days, regardless of whether a fall had occurred.
At each visit, participants contributed three complementary categories of data. 
First, \textbf{structured survey-based assessments} captured physical, cognitive, and functional fall risk factors using validated instruments (e.g., the 7-item Short Fall Efficacy Scale), resulting in low-frequency, visit-level tabular measurements. 
Second, \textbf{clinical measurements}, including body composition, grip strength, and standardized evaluations of muscle strength and balance, provided objective physiological measures at each visit. 
Third, to characterize physical activity in naturalistic settings, participants wore ActiGraph triaxial accelerometers for seven consecutive days following each visit, generating high-frequency, minute-level \textbf{time-series physical activity data} such as step counts, vector magnitude (VM), and posture states (e.g., sitting, standing, and lying).
\textit{\textbf{Fall incidence}} served as the primary outcome and was recorded as self-reported, time-stamped events, resulting in sparse and irregular event sequences for most participants. No participant in the analyzed cohort died before the end of follow-up; therefore, death-related competing-risk censoring did not arise in the current experiments.
Together, these multimodal data spanning survey-based, clinical, and wearable-derived measurements across multiple temporal resolutions enable the identification of personalized and dynamically evolving fall risk factors, supporting adaptive fall risk identification and prevention.

\section{PAFIR Framework}
The \textbf{P}ersonalized and \textbf{A}daptive \textbf{F}eature selection framework for fall risk \textbf{I}dentification and p\textbf{R}evention (PAFIR; Fig.~\ref{fig:overview}) is designed to tackle the challenges of selecting effective fall risk factors from longitudinal, multimodal health data and developing personalized prevention strategies for healthcare applications. PAFIR integrates three key innovations: (1) hierarchical structure linking feature groups to individual features; (2) a comparison-driven feature selection mechanism that iteratively contrasts candidate and reference feature groups to refine the identification of clinically relevant features; and (3) a sparse-aware reward design with clinical proxies.PAFIR optimizes RL policies for outcome-relevant feature selection, enhancing the interpretability and effectiveness of longitudinal health modeling in real-world, dynamic care environments. Formally, let $i$ index participants and let $\mathcal{X}$ denote the complete feature space, including structured clinical variables and wearable-derived temporal features. PAFIR learns a feature-selection policy shared across the training population and applies it to each participant-specific state to produce an individualized selected feature set. The index $t$ denotes an asynchronous information update. Let $\mathbf{s}_{i,t}$ denote the multimodal state available for participant $i$ at update $t$, and let $F_{i,t-1}\subseteq\mathcal{X}$ denote the active feature set carried into that update. The three coordinated agent policies define a distribution over the composite feature-selection action, $\pi_{\theta}(a_{i,t}\mid\mathbf{s}_{i,t},F_{i,t-1})$, where $a_{i,t}$ consists of candidate selection, reference selection, and an add, remove, or retain operation. Applying $a_{i,t}$ produces an updated feature set $F_{i,t}$, which may contain more, fewer, or the same number of features as $F_{i,t-1}$. At each update, $y_{i,t}$ denotes the available outcome signal: the observed fall outcome for a fall-related update, or a visit-level clinical proxy, such as FES-I or BBS, for a scheduled assessment update.

\subsection{Hierarchical Multimodal Structure for Fall Risk Feature Selection}

To effectively model both structural clinical assessments and temporal dynamics in time-series wearable sensor data, we construct a state representation $\mathbf{s}_{i,t}$ that integrates structured clinical features and time-series physical activity signals. This design captures both the organization of clinical variables and the evolution of patient status across visits.

\paragraph{Hierarchical Structural Representation}

Longitudinal clinical data and time-series physical activity signals form a structured and multimodal representation of patient health. Features are organized into clinically meaningful feature groups based on their functional roles. For example, the body composition group includes variables such as BMI and weight, the physical functional group includes the measurements of grip strength, and the psychological group includes assessments such as the FES, which captures fear of falling. Features within each group are often correlated and reflect related aspects of participant health, while relationships across groups capture interactions among broader fall-risk domains. Deterioration in physical performance often co-occurs with elevated fear of falling, and cognitive decline frequently accompanies functional limitation. Capturing both levels is essential for identifying which risk factors are clinically relevant and how they jointly evolve over time. We partition the feature space $\mathcal{X}$ into $M$ clinically meaningful feature sets, $\{\mathcal{V}_1,\mathcal{V}_2,\dots,\mathcal{V}_M\}$, where each $\mathcal{V}_m\subseteq\mathcal{X}$ consists of a subset of related features, and $\bigcup_{m=1}^{M}\mathcal{V}_m=\mathcal{X}$.

Within each clinically defined group, features are represented as nodes in an intra-group graph $\mathcal{G}_m=(\mathcal{V}_m,\mathcal{E}_m)$. The feature groups are predefined according to the PEER assessment domains, and every pair of features within the same domain is connected, forming a fully connected intra-group graph. This construction encodes shared clinical-domain membership and enables related features to be evaluated jointly.

Across groups, we further define an inter-group topology $\mathcal{T} = (\mathcal{U}, \mathcal{W})$, where each node $u_m \in \mathcal{U}$ represents a feature group $\mathcal{V}_m$.
For participant $i$ at update $t$, let $\Phi_{i,t}^{(m)} = \sum_{v\in \mathcal{V}_m}\phi_{i,t,v}$ denote the initial importance score of group $m$, where $\phi_{i,t,v}$ denotes the importance of feature $v$. These initial importance scores are used to construct the initial structural representation and to initialize the reward reference and feature-selection process.
Each weighted edge $w_{mn} \in \mathcal{W}$ reflects the co-variation in initial importance between groups $m$ and $n$, that is, the extent to which
their aggregated initial importance scores increase or decrease together across participants and updates, 
$w_{mn} = \operatorname{Corr}_{(i,t)\in\mathcal{D}_{\mathrm{train}}}\!\left(\Phi_{i,t}^{(m)}, \Phi_{i,t}^{(n)}\right)$, where the correlation is computed across available participant-update observations in the training data.

For each feature $v \in \mathcal{V}_m$, we construct a structural representation $\mathbf{h}_{i,t,v}$ that captures its relationships with other features in the same clinical domain for participant $i$ at update $t$. 
Let $\mathbf{H}_{i,t} = \left\{\mathbf{h}_{i,t,v}: v\in\mathcal{V}_m,\; m=1,\ldots,M \right\}$ denote the collection of structural feature representations at update $t$. These representations summarize within-group dependencies and group-level organization, providing a compact encoding of the structured feature space. Detailed descriptions of the encoding procedure are provided in Appendix~\ref{appendix:encoding}.

\paragraph{Temporal Representation} 
In addition to structured clinical features, physical activity data are inherently time-series observations, capturing fine-grained behavioral dynamics across time. While clinical variables are typically recorded at discrete visits, physical activity signals provide continuous measurements between visits, offering complementary information about patient behavior. Let $\mathbf{X}_{i,:,b}^{(\tau)} \in \mathbb{R}^{L \times 1}$ denote the time-series data collected for participant $i$ following visit $\tau$, where $L$ is the sequence length and $b$ denotes the corresponding temporal feature. At the corresponding update $t$, each time series is mapped to a latent representation $\mathbf{z}_{i,t,b}$ that captures temporal patterns such as step counts and physical activity vector magnitude. A detailed description of the construction of the temporal representation $\mathbf{z}_{i,t,b}$ is provided in Appendix~\ref{appendix:timeencoding}.

\paragraph{State Construction}
Let $\mathbf{Z}_{i,t}= \left\{\mathbf{z}_{i,t,b}: b\in\mathcal{B} \right\}$ denote the set of temporal embeddings available for participant $i$ at update $t$, where $\mathcal{B}$ denotes the set of time-series physical activity features.
The final state representation is constructed by integrating structural and temporal information,
\begin{align}
 \mathbf{s}_{i,t} = f\left(\mathbf{H}_{i,t}, \mathbf{Z}_{i,t}\right),
\end{align}
where $f(\cdot)$ represents a fusion mechanism that aligns structured features with temporal dynamics. This formulation allows $\mathbf{s}_{i,t}$ to capture both the topology of the feature space and the temporal evolution of patient behavior across visits, supporting more stable and context-aware feature selection.

\subsection{Adaptive Comparison-Driven Feature Selection Mechanism}
We design an adaptive comparison-driven feature selection mechanism to iteratively identify outcome-relevant features under complex and correlated clinical settings. The key idea is to decompose feature selection into a sequence of structured decisions that balance candidate discovery, comparative evaluation, and refinement~\citep{zhong2012correlation}. To implement this process, we adopt a three-agent feature selection framework~\citep{ying2024topology}, where each agent is responsible for a distinct role in the stepwise selection procedure. 
Specifically, the candidate agent proposes a feature group for evaluation, the reference agent selects a competing group for comparison, and the operation agent determines whether the active feature set should be updated based on their relative outcome relevance (in Fig.\ref{fig:overview}). A single agent would need to select the candidate group, reference group, and operation jointly. The three-agent formulation instead factorizes this action into smaller conditional decisions. The agents are executed sequentially, share the same participant-specific state and reward, and condition each downstream decision on the preceding actions. They therefore form a coordinated feature-selection policy rather than independent decision makers.

Let $F_{i,t-1}\subseteq\mathcal{X}$ denote the active feature set for participant $i$ before update $t$, and let $\mathcal{C}= \{\mathcal{V}_1,\mathcal{V}_2,\ldots,\mathcal{V}_M\}$ denote the collection of feature groups derived from the hierarchical structure. At each update, the candidate agent selects a feature group based on its potential relevance to the outcome, $C_{i,t}^{\mathrm{cand}} \sim \pi_{\theta_1} \left(C \mid \mathbf{s}_{i,t}, F_{i,t-1} \right), \quad C_{i,t}^{\mathrm{cand}}\in\mathcal{C}$. Rather than evaluating the candidate group in isolation, the reference agent selects a comparison group from the remaining feature space, $C_{i,t}^{\mathrm{ref}} \sim \pi_{\theta_2} \left(C \mid \mathbf{s}_{i,t}, F_{i,t-1}, C_{i,t}^{\mathrm{cand}}\right)$, $C_{i,t}^{\mathrm{ref}}\in
\mathcal{C}\setminus\{C_{i,t}^{\mathrm{cand}}\}$. The operation agent then selects an add, remove, or retain operation, $o_{i,t} \sim \pi_{\theta_3} \left(o \mid \mathbf{s}_{i,t}, F_{i,t-1}, C_{i,t}^{\mathrm{cand}}, C_{i,t}^{\mathrm{ref}} \right)$, where $o_{i,t} \in \{\mathrm{add},\mathrm{remove},\mathrm{retain}\}$. The resulting composite action is $a_{i,t} = \left( C_{i,t}^{\mathrm{cand}}, C_{i,t}^{\mathrm{ref}}, o_{i,t} \right)$.

The candidate and reference groups are evaluated relative to the current active feature set using the available outcome feedback. The same comparison procedure is subsequently applied within each retained group to refine the selection at the individual-feature level while preserving the hierarchical structure.

\subsection{Sparse-Aware Reward Design With Clinical Proxies}

Learning an effective feature selection policy from fall incidence is challenging due to the sparsity and temporal irregularity of observed events. In many cases, participants do not experience a fall across multiple visits, resulting in limited and delayed feedback if rewards are defined solely based on fall occurrence. To address this, we design a dynamic reward mechanism that provides continuous and adaptive feedback throughout the selection process. At the beginning of training, baseline feature-importance scores are used to provide an informed initial reward reference and selection prior. During training, the reward is computed using the outcome information available at each update. In addition to fall incidence, the reward signal incorporates clinically validated proxy measures that are available at every visit and are known to be associated with fall risk, such as the Falls Efficacy Scale-International (FES-I,~\cite{yardley2005development}) and the BTrackS Balance Tracking System score (BBS,~\cite{levy2018validity}). These proxy signals, as fall risk appraisal, provide continuous supervision even in the absence of observed falls, allowing the model to receive meaningful feedback at every step~\citep{thiamwong2020assessing}. 

Specifically, when a fall incident occurs between visits (e.g., between Visit 2 and Visit 3), the reward is immediately adjusted to reflect the contribution of the selected features. Feature groups that are more strongly associated with fall risk receive higher rewards, while those that do not contribute are penalized. At scheduled assessment updates without fall feedback, the reward is evaluated using an available clinical proxy, such as FES-I or BBS. This design ensures that rare but clinically important fall events provide strong learning signals when they occur, while proxy-based signals maintain stable feedback throughout the remaining visits. The reward function is defined as
\begin{equation}
\label{reward}
r_{i,t}
= \operatorname{Perf}\left(F_{i,t},y_{i,t}\right) - p_{i,t-1},
\end{equation}
where $\operatorname{Perf}(F_{i,t},y_{i,t})$ denotes the evaluation measure used for the fall outcome when fall feedback is available and for the clinical proxy outcome otherwise. The term $p_{i,t-1}$ denotes the best previously observed performance for the corresponding type of outcome feedback, following \citet{zhong2012correlation}.

Each agent is trained using a separate deep Q-network. For agent $j\in\{\mathrm{cand},\mathrm{ref},\mathrm{op}\}$, the temporal-difference loss at update $t$ is,
\begin{align}
\mathcal{L}_{i,t}^{(j)}
=
\Bigg(
Q_j\left(\mathbf{s}_{i,t},a_{i,t}^{(j)}\right)
-
\bigg[
r_{i,t}
+
\gamma
\max_{a'}
Q_j\left(\mathbf{s}_{i,t+1},a'\right)
\bigg]
\Bigg)^2,
\end{align}
where $Q_j$ denotes the action-value function of agent $j$, $a_{i,t}^{(j)}$ denotes the corresponding agent action, $r_{i,t}$ is the shared reward, and $\gamma$ is the discount factor. All Q-networks are randomly initialized and trained using the temporal-difference loss. The baseline feature importance scores are used to initialize the reward and the initial feature selection.

This reward design allows the model to learn from both sparse fall events and continuous clinical signals, making the learning process more stable while maintaining alignment with clinically meaningful patterns.

\section{In-Field Experiment Evaluations}
\noindent\textbf{PEER Fall Prevention Data.} We evaluate PAFIR using longitudinal, multimodal data from the PEER study collected in real-world settings, emphasizing adaptive feature selection across repeated clinical visits. The dataset includes 341 community-dwelling participants assessed at four time points, yielding 1,364 participant–visit observations. Following each scheduled visit, participants wore sensors continuously for seven consecutive days, producing minute-level wearable data and totaling over 2.5 million time-stamped observations across all participants and visits.

At each scheduled visit, participants contribute high-dimensional structured clinical assessments, spanning physical, cognitive, psychological, and functional domains. After preprocessing, the structured modality comprises 587 fall-related features organized into 35 clinically meaningful feature groups, including body composition, physical activity, cognitive function, balance confidence, and psychological status. Each participant–visit instance is modeled as a graph-structured input, which is encoded by a graph encoder to capture hierarchical and correlational dependencies among risk factors.

In parallel, minute-level wearable sensor data, including step counts, vector magnitude (VM), and posture states (sitting, standing, and lying), are processed by a time-series encoder to model temporal dynamics within each seven-day monitoring period. The resulting temporal representations are aligned with visit-level structured embeddings and fused via a cross-attention Transformer module, yielding a hierarchical latent state that integrates structural and temporal information and serves as the input to the reinforcement learning component.

Fall incidence events are recorded as self-reported, time-stamped outcomes, providing sparse and delayed supervision signals that guide adaptive learning of feature relevance across visits. The proxy outcomes, namely the FES-I and BBS scores, are assessed at each clinical visit.

\noindent\textbf{PAFIR Implementation.} Reinforcement learning feature selection operates at the visit level or when a fall incidence occurs, enabling the model to update feature importance dynamically as new temporal and structural information becomes available.
The structured feature space is encoded using a six-layer graph encoder with a hidden dimension of 128, residual connections, and a dropout rate of 0.5.
Mean pooling and Laplacian positional encoding with up to ten eigenvectors are applied to aggregate node-level representations.
The temporal encoder uses four stacked Transformer encoder blocks, each with four attention heads and a feed-forward dimension of 256, with wearable inputs projected through a shared linear embedding layer.

The model is trained for 100 epochs using the Adam optimizer with a learning rate of 0.001. The Q-network of each agent is optimized using the temporal-difference loss defined in the previous section. Binary cross-entropy loss is used for the fall-outcome classification model employed in the fall-based reward calculation. 
All experiments are conducted on NVIDIA Quadro RTX 6000 GPUs using CUDA 13.0. Each random-seed run requires approximately 101 seconds, and the complete 20-seed longitudinal stability analysis requires approximately 34 minutes. The peak allocated GPU memory is approximately 240 MB. Computational resource requirements are reported in Appendix~\ref{appendix:computational_efficiency} and Table~\ref{tab:runtime}.
\vspace{-0.5em}
\subsection{Results: Recommendations for Fall Prevention in PEER Study}
\paragraph{Community-Level Recommendations for Fall Prevention.}
\begin{figure}[ht]
    \centering
    \subfigure[PEER intervention site: \textit{Kinneret}]{
        \includegraphics[width=0.45\linewidth]{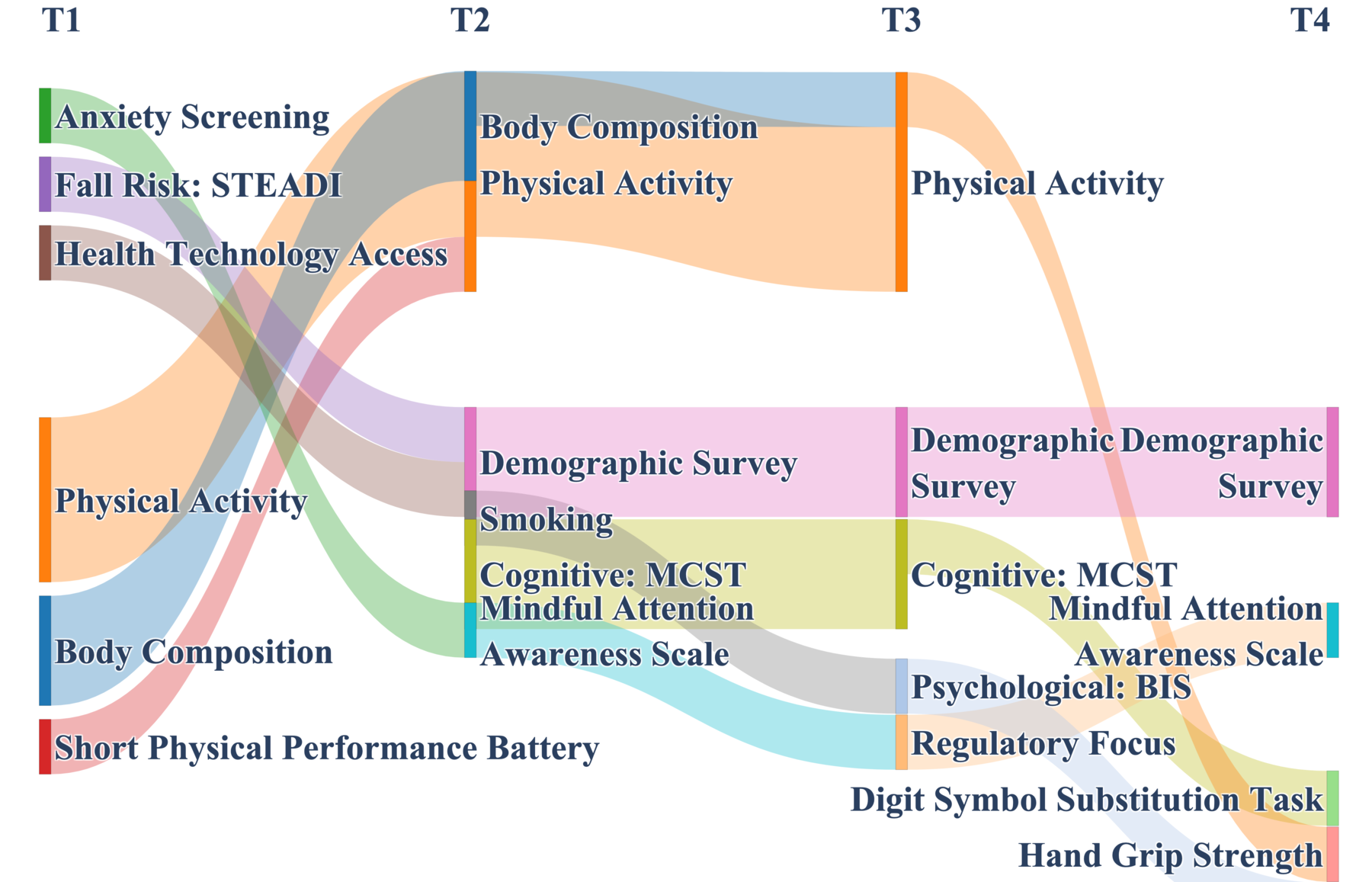}
        \label{subfig:Kinneret}
    } 
    \subfigure[Control site: \textit{LCA}]{
        \includegraphics[width=0.45\linewidth]{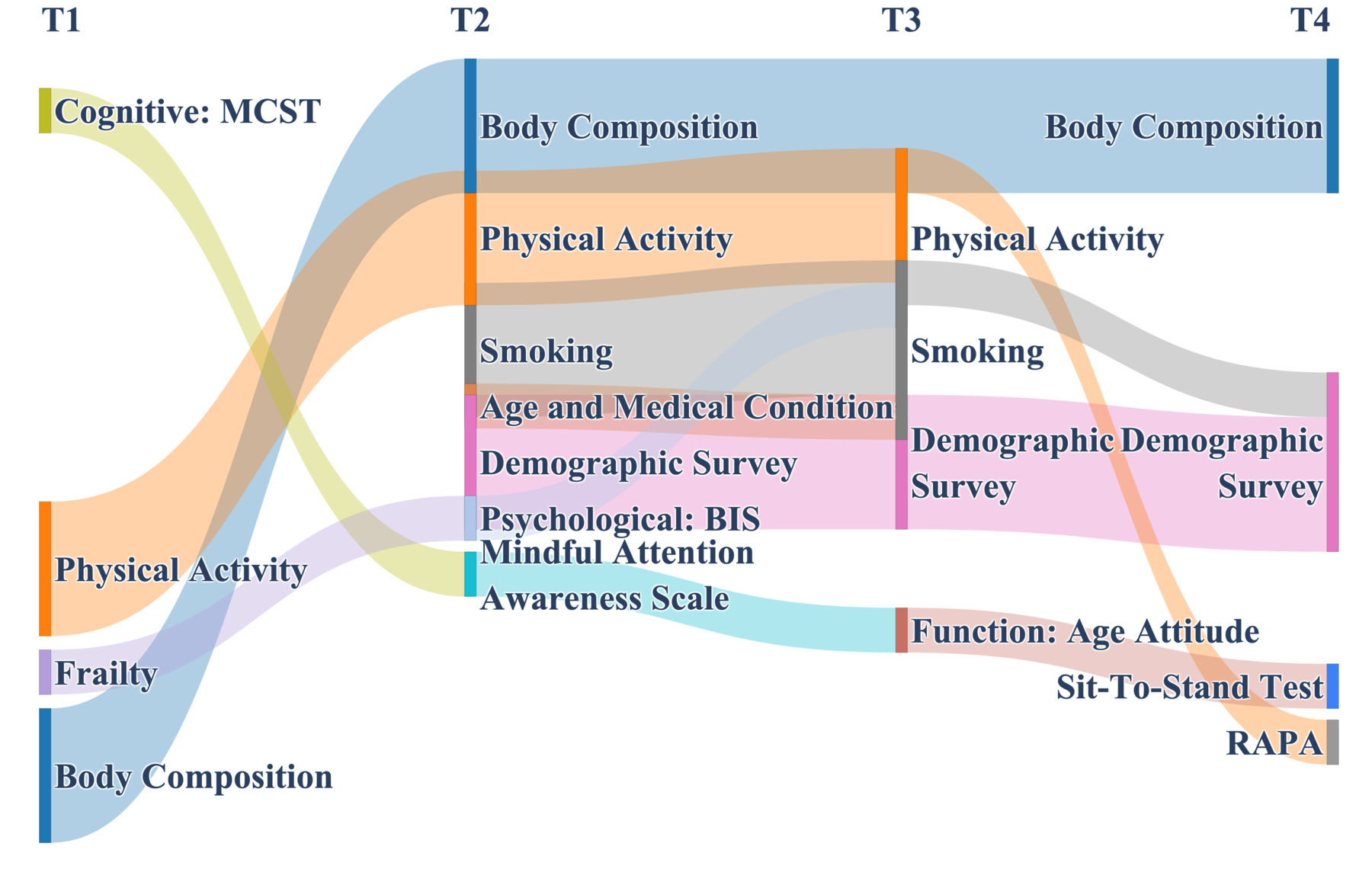}
        \label{subfig:LCA1}
    }
    \caption{\small Top fall risk factors identified across four visits at the community level.
    }
    \label{fig:site_feature}
\end{figure}
We first describe how PAFIR produces the longitudinal community-level results shown in Fig.~\ref{fig:site_feature}.
PAFIR is applied to multimodal data collected at four visits (T1--T4), including clinical assessments, survey-based measures, and wearable-derived physical activity features.
The model is trained using data from all participants at the community level.
At each visit, PAFIR selects important fall risk factors by considering both changes over time and relationships between feature groups.
This procedure produces visit-level feature selection results for each site, allowing us to examine how selected fall risk factors evolve over time in both the intervention and control communities.

Across both sites, PAFIR consistently identifies \textit{physical activity}, \textit{body composition}, and \textit{demographic information} across visits, which aligns with established evidence that mobility-related measures and body composition are among the most persistent predictors of falls in community-dwelling older adults~\citep{thiamwong2023body, kohler2025multifactorial}. At the Kinneret intervention site (Fig.~\ref{fig:site_feature}\subref{subfig:Kinneret}), PAFIR identifies a broader set of fall risk features beyond physical performance, including psychological and behavioral measures such as \textit{anxiety screening}, \textit{mindful attention awareness}, \textit{regulatory focus}, and \textit{psychological inhibition sensitivity} (BIS), which are selected across multiple visits. This pattern is consistent with prior evidence that psychological factors, including fear of falling, attentional control, and behavioral regulation, are independently associated with fall risk in community-dwelling older adults~\citep{yi2022relationship, sturnieks2025cognitive}. The consistent selection of psychological and behavioral features at Kinneret across all four visits suggests that PEER's cognitive and psychological components actively engage these domains as modifiable risk factors~\citep{Thiamwong2023, thiamwong2020physio}, making them detectable and trackable by PAFIR throughout the intervention period. This is in line with evidence that multicomponent interventions combining physical exercise with cognitive-behavioral elements produce greater reductions in fall risk than single-component physical programs alone~\citep{liu2025effectiveness}.  In contrast, the LCA control site (Fig.~\ref{fig:site_feature}\subref{subfig:LCA1}) predominantly selects traditional physical and functional risk factors, including \textit{frailty}, \textit{physical activity}, \textit{body composition}, \textit{age-related medical conditions}, and \textit{functional performance measures} (Sit-to-Stand, RAPA), across visits. This pattern reflects the well-documented role of physical frailty, muscle function, and chronic health conditions as core fall risk determinants in older adult populations not receiving structured behavioral interventions~\citep{chittrakul2020physical, saunders2025risk}. The absence of psychological and behavioral features at LCA reflects the profile expected in a population not receiving structured multidimensional intervention, where conventional physical and functional factors dominate the fall risk landscape~\citep{liu2025diffusion, nguyen2024unveiling}. The divergence in feature profiles between the two sites provides indirect evidence of PEER's effectiveness, demonstrating that an effective intervention not only addresses physical fall risk but also renders psychological and behavioral risk factors consistently detectable over time~\citep{liu2025effectiveness}. These findings reinforce the clinical value of integrating psychological engagement and cognitive reframing into community-level fall prevention, complementing physical activity and body composition monitoring as a more comprehensive risk management strategy.

\vspace{-0.1in}

\paragraph{Temporal Evolving of Fall Risk Factors}
\begin{figure}[t]
    \centering
    \includegraphics[width=\linewidth]{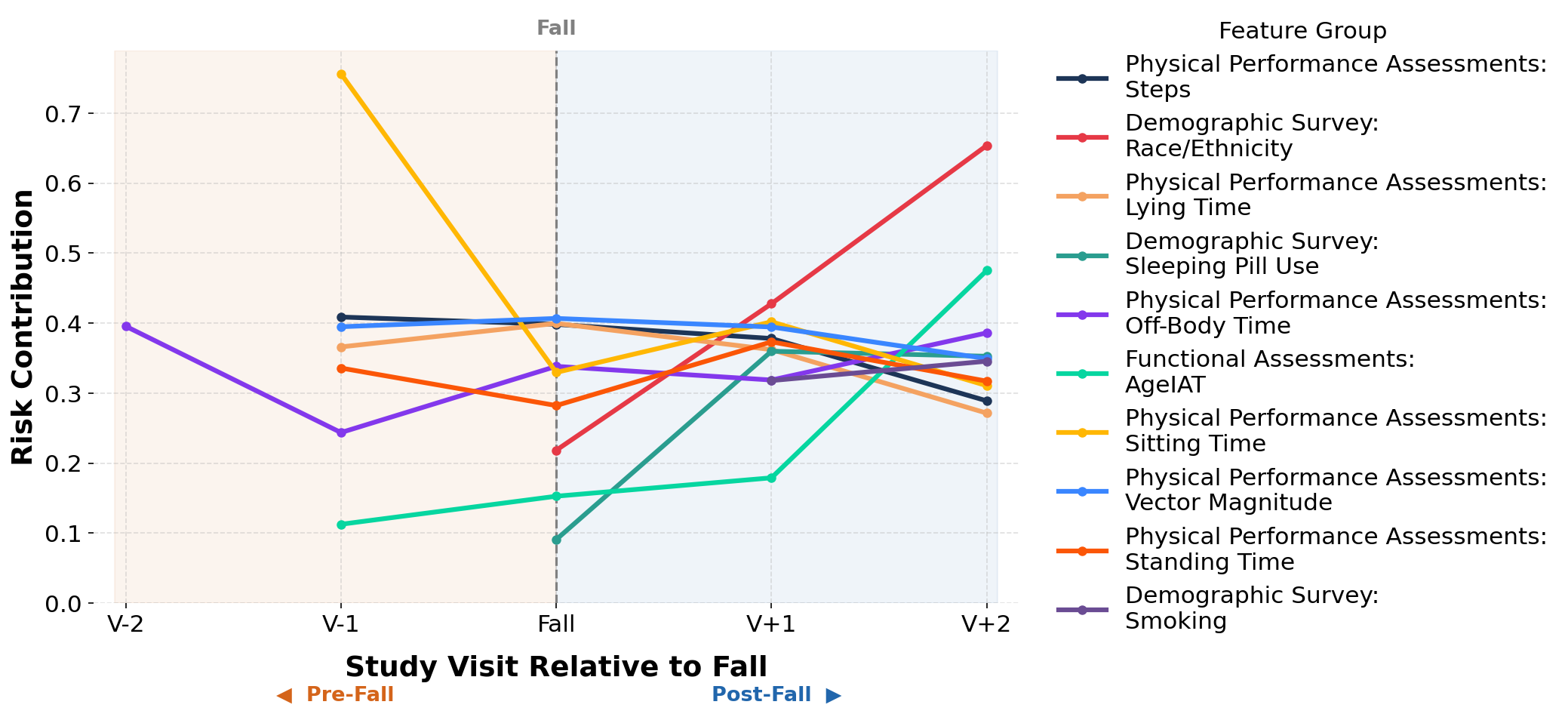}
    \caption{\small Population-level fall risk contributions across study visits relative to a fall event. Visits are aligned by their order before and after the fall, and the vertical dashed line indicates the fall event. Markers at the dashed line denote the fall-related update, whereas the remaining markers denote scheduled visit-level contributions.}
    \label{fig:temporal_evolution_fall}
\end{figure}
We next examine how model-derived feature contributions vary across visits relative to a recorded fall event. Participant trajectories are aligned according to the visit order before and after the fall, with V$-2$ and V$-1$ representing pre-fall visits and V$+1$ and V$+2$ representing post-fall visits. As shown in Fig.~\ref{fig:temporal_evolution_fall}, several features exhibit distinct temporal patterns. Activity-related features, including \textit{Steps}, \textit{Vector Magnitude}, \textit{Lying Time}, \textit{Off-Body Time}, and \textit{Sitting Time}, receive relatively high contributions before or around the fall, consistent with the established importance of mobility and activity patterns in fall-risk characterization~\citep{Thiamwong2023,taheri2025fear}. In contrast, the contributions of \textit{Race/Ethnicity}, \textit{Sleeping Pill Use}, \textit{AgeIAT}, and \textit{Smoking} increase across later visits, suggesting that demographic, behavioral, and functional factors remain relevant during post-fall follow-up~\citep{xu2019body,colon2024risk,thiamwong2023body}. The increasing contribution of \textit{AgeIAT} may also reflect the broader role of cognitive and psychological factors in fall-related outcomes~\citep{sturnieks2025cognitive}. These patterns represent changes in the relevance assigned to each feature by PAFIR rather than changes in the underlying raw feature values, illustrating how the framework updates feature priorities across longitudinal assessments.

\paragraph{Personalized Recommendation for Fall Prevention}
\begin{figure}[h]
    \centering
    \includegraphics[width=\linewidth]{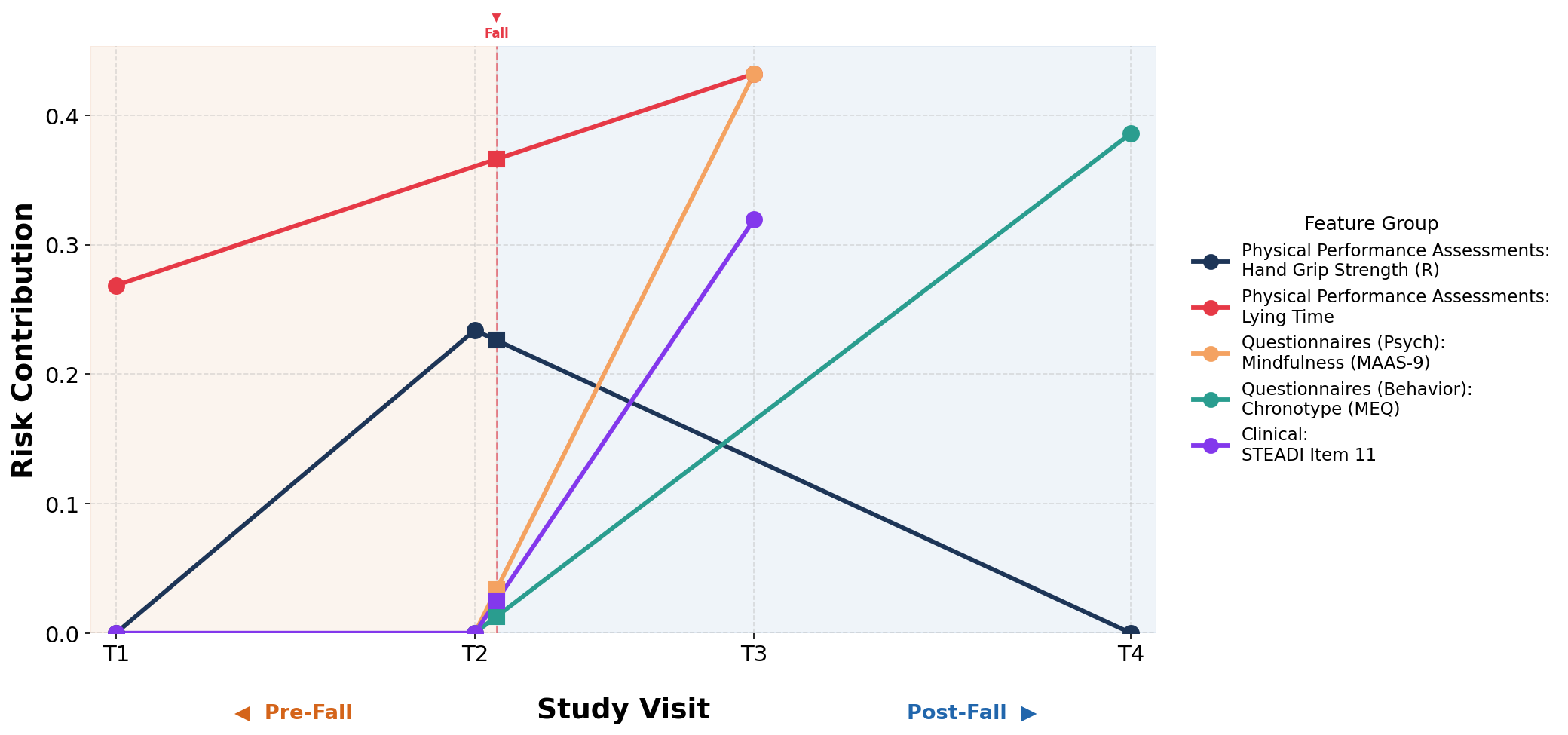}
    \caption{\small Personalized Temporal Evolution of Feature Importance Before and After Fall for \textit{Subject 1017}. The vertical dashed line indicates the recorded fall event. Circles denote scheduled visit-level contributions, whereas squares denote contributions at the fall-related update.}\label{fig:personalized_recommendation}
\end{figure}
We next present a personalized analysis to illustrate how PAFIR supports individual-level longitudinal monitoring through temporally evolving and adaptive feature selection. Figure~\ref{fig:personalized_recommendation} visualizes the trajectories of selected fall-related features for Subject~1017 across four study visits, with a recorded fall occurring shortly after the T2 visit.
Figure~\ref{fig:personalized_recommendation} reveals a clear transition in feature contributions across the pre- and post-fall periods. At T1, \textit{Lying Time} (Physical Performance Assessments) exhibits the highest contribution, indicating that sedentary behavior is a prominent fall-related feature at baseline~\citep{xu2019body,Thiamwong2023}. At T2, shortly before the fall, \textit{Hand Grip Strength (R)} (Physical Performance Assessments) receives the highest visit-level contribution, whereas \textit{Lying Time} is not selected at the scheduled visit.
This pre-fall pattern, centered on physical activity and muscle function, is consistent with established fall-risk factors among older adults~\citep{thiamwong2023body}.  
Following the fall, the selected feature profile broadens across multiple domains. At the fall-related update shortly after T2, \textit{Lying Time} receives the largest contribution, while \textit{Hand Grip Strength} remains prominent. At T3, \textit{Lying Time} reaches its highest observed contribution, alongside \textit{Mindfulness (MAAS-9)}(Questionnaires--Psychological), while \textit{STEADI Item 11} (Clinical) is also selected. At T4, \textit{Chronotype (MEQ)} (Questionnaires--Behavior) becomes the most prominent selected feature~\citep{yi2022relationship,taheri2025fear}. The increased contribution of \textit{Lying Time} after the fall may be consistent with reduced mobility or fear-related activity restriction, while the emergence of psychological, behavioral, and clinical screening features illustrates the multidimensional nature of post-fall follow-up. 
From a personalized monitoring perspective, these temporal patterns demonstrate PAFIR's capacity to generate adaptive, individual-level feature priorities. The pre-fall selections emphasize physical activity and muscle function, whereas the post-fall profile expands to include physical, psychological, behavioral, and clinical screening domains, suggesting that these areas may warrant broader review during subsequent follow-up~\citep{thiamwong2023body}.

\subsection{Benchmark and Ablation Analysis}
\noindent\textbf{Evaluation Metrics.} 
We compare PAFIR with several state-of-the-art feature selection baselines, including TAR~\citep{ying2024topology}, SAT~\citep{chen2022structure}, TTG~\citep{khurana2018feature}, PCA~\citep{mackiewicz1993principal}, Group LASSO~\citep{yuan2006model}, and LASSO~\citep{tibshirani1996regression}. Participants are randomly divided into 80\% training and 20\% testing sets, and all longitudinal observations from the same participant are retained in the same partition. All data-dependent quantities, including feature-importance scores and inter-group correlations, are estimated using the training partition only. All experiments are repeated 20 times with different random seeds, and the results are reported as mean $\pm$ standard deviation. We assess the quality of feature selection directly using four complementary metrics. 
\textbf{Feature Recovery Rate (FRR) }measures the proportion of fall-relevant features identified. \textbf{False Discovery Rate (FDR)} quantifies the proportion of irrelevant features selected~\citep{benjamini1995controlling}. \textbf{Selection Stability} evaluates consistency across runs~\citep{kuncheva2007stability}. \textbf{Correlation Recovery} assesses how well feature dependencies are preserved~\citep{meinshausen2010stability, meinshausen2006high}.
The complete feature- and group-level initial reference sets used for evaluation are provided in Appendix~\ref{appendix:reference_sets}.

\begin{table*}[t]
\centering
\caption{Performance results for fall risk factor selection on the PEER study.}
\label{tab:fall_risk_overall}
\resizebox{\textwidth}{!}
{\small
\begin{tabular}{c|cccc}
\toprule
\textbf{Method} 
& \textbf{FRR} $\uparrow$ 
& \textbf{FDR} $\downarrow$ 
& \textbf{Selection Stability} $\uparrow$ 
& \textbf{Correlation Recovery} $\uparrow$ \\
\midrule
\textbf{PAFIR (Ours)}
& \textbf{0.874 $\pm$ 0.007}
& \textbf{0.207 $\pm$ 0.009}
& \textbf{0.990 $\pm$ 0.003}
& \textbf{0.974 $\pm$ 0.014}
\\
TAR~(\citeyear{ying2024topology})
& 0.795 $\pm$ 0.009
& 0.209 $\pm$ 0.008
& 0.989 $\pm$ 0.001
& 0.973 $\pm$ 0.014
\\
SAT~(\citeyear{chen2022structure})
& 0.731 $\pm$ 0.010
& 0.311 $\pm$ 0.011
& 0.985 $\pm$ 0.003
& 0.973 $\pm$ 0.013
\\
TTG~(\citeyear{khurana2018feature})
& 0.664 $\pm$ 0.009
& 0.379 $\pm$ 0.012
& 0.985 $\pm$ 0.004
& 0.972 $\pm$ 0.013
\\
PCA~(\citeyear{mackiewicz1993principal})
& 0.624 $\pm$ 0.015
& 0.434 $\pm$ 0.012
& 0.978 $\pm$ 0.002
& 0.971 $\pm$ 0.016
\\
Group LASSO~(\citeyear{yuan2006model})
& 0.573 $\pm$ 0.154                                                   
& 0.667 $\pm$ 0.061                                                   
& 0.879 $\pm$ 0.134                                    
& 0.970 $\pm$ 0.005
\\
LASSO~(\citeyear{tibshirani1996regression})
& 0.278 $\pm$ 0.005
& 0.500 $\pm$ 0.000
& 0.990 $\pm$ 0.009
& 0.902 $\pm$ 0.011
\\
\bottomrule
\end{tabular}
}
\vspace{-0.08in}
\end{table*}

\paragraph{Performance and Ablation Study}
Table~\ref{tab:fall_risk_overall} summarizes the overall feature selection performance on the PEER study. PAFIR achieves the highest Feature Recovery Rate (FRR) and Selection Stability, while maintaining a competitive False Discovery Rate (FDR). This indicates that PAFIR is able to consistently identify clinically relevant fall-risk features. Its advantage is especially clear in Correlation Recovery, where PAFIR better preserves the relationships among features compared to other methods. This is important in practice, as it allows clinicians to interpret how different risk factors interact, rather than treating them as independent variables.

\begin{table*}[t]
\centering
\caption{Ablation Study for Fall Risk Factors Selection on the PEER Study.}
\label{tab:ablation_fall_risk}
\resizebox{\textwidth}{!}
{\small
\begin{tabular}{c|cccc}
\toprule
\textbf{Method} 
& \textbf{FRR} $\uparrow$ 
& \textbf{FDR} $\downarrow$ 
& \textbf{Selection Stability} $\uparrow$ 
& \textbf{Correlation Recovery} $\uparrow$ \\
\midrule
PAFIR (Full)
& \textbf{0.874 $\pm$ 0.007}
& \textbf{0.207 $\pm$ 0.009}
& \textbf{0.990 $\pm$ 0.003}
& \textbf{0.974 $\pm$ 0.014}
 \\
 w/ Random Feature-Importance Prior
& 0.870 $\pm$ 0.015 
& 0.211 $\pm$ 0.008 
& 0.980 $\pm$ 0.011 
& 0.970 $\pm$ 0.014 
\\
w/o Hierarchical Structure 
& 0.768 $\pm$ 0.010 
& 0.229 $\pm$ 0.009 
& 0.769 $\pm$ 0.010 
& 0.752 $\pm$ 0.008  
\\
w/o Three-Agent Comparison-Driven Mechanism
& 0.678 $\pm$ 0.005 
& 0.327 $\pm$ 0.013 
& 0.675 $\pm$ 0.007 
& 0.765 $\pm$ 0.033 
\\
w/o Sparse-Aware Rewards
& 0.655 $\pm$ 0.011 
& 0.325 $\pm$ 0.008 
& 0.665 $\pm$ 0.007 
& 0.701 $\pm$ 0.013 
\\
\bottomrule
\end{tabular}
}
\vspace{-0.1in}
\end{table*}

Table~\ref{tab:ablation_fall_risk} shows that removing any component leads to consistent performance degradation. Without the hierarchical structure, FRR and selection stability decrease. Without the comparison-driven selection mechanism, FRR decreases and FDR increases. Without the sparse-aware reward, FRR and correlation recovery are the lowest. These results highlight the complementary roles of structural representation, comparison-driven selection, and reward design.
Replacing the baseline feature-importance prior with a random prior yields only modest performance changes, indicating that the prior provides useful guidance without determining the learned policy. Additional results are reported in Appendix~\ref{appendix:feature_group_results}, Table~\ref{tab:fall_risk_node_group}, and Appendices~\ref{appedix:benchmark}--\ref{appendix:generality}. The public benchmark experiments support methodological applicability but do not constitute external clinical validation.

\vspace{-0.05in}
\paragraph{Longitudinal Stability in the No-Fall Subgroup} Among participants with no recorded falls, PAFIR shows high group-level stability across visits, with an overall Jaccard similarity of $0.890 \pm 0.009$. In contrast, individual-feature selections vary more substantially. Detailed group- and feature-level results are provided in Appendix~\ref{appendix:no_fall_stability} and Table~\ref{tab:no_fall_stability}.

\vspace{-1em}
\section{Conclusion, Future Work, and Limitations}
This study presents PAFIR, a multi-level RL-based feature selection framework designed to support personalized and adaptive fall prevention for older adults. By modeling the hierarchical structural assessments and aligning evolving physical activities from wearable sensors, PAFIR adaptively identifies temporally dynamic and fall risk factors, which can inform personalized recommendations. Particularly, PAFIR demonstrates its potential to identify the actionable early warning signal from the in-field PEER study and to inform personalized monitoring and fall-prevention planning as risk factors evolve. 

\paragraph{Future Work and Limitations}
Future directions for PAFIR include integration with digital health and mHealth platforms, such as Ecological Momentary Assessment (EMA) systems, to enable real-time monitoring and adaptive feedback. By incorporating high-frequency self-reported data on symptoms and behaviors, PAFIR may support just-in-time adaptive interventions (JITAIs) that deliver personalized prompts when relevant changes in mobility or functional capacity are identified. This integration could bridge passive sensing with active behavioral support. Future studies will evaluate PAFIR-powered digital interventions in community-dwelling older adults. External validation is currently constrained by the lack of comparable longitudinal multimodal fall-prevention cohorts.


\acks{
We thank the reviewers and the Area Chair for their constructive feedback. This research was supported by NIH the National Institute on Minority Health and Health Disparities (R01MD018025) and the NIH Office of the Director, Chief Officer for Scientific Workforce Diversity (COSWD) (3R01MD018025-02S1), and the Learning Institute for Elders at University of Central Florida Richard Tucker Gerontology Applied Research Grant.
}

\bibliography{PAFIR_MLHC2026}

\clearpage
\appendix

\section{Supplement Overview}
This supplementary document presents additional information to support the main manuscript. It includes detailed descriptions of the datasets used, the full configuration of the PAFIR model, the training and evaluation procedures, and extended experimental results. These materials aim to enhance the reproducibility and transparency of our study and provide deeper insights into the implementation and performance of PAFIR across different settings. The code implementing the proposed PAFIR framework and all experimental pipelines is available at: \url{https://github.com/changliu1993-cl/PAFIR}

\section{In-Field Evaluation: PEER Study on Fall Prevention in Older Adults}

\subsection{PEER Intervention and Control Across Visits}

\begin{figure}[ht]
\centering
\subfigure[PEER Intervention]{
\includegraphics[width=0.45\linewidth]{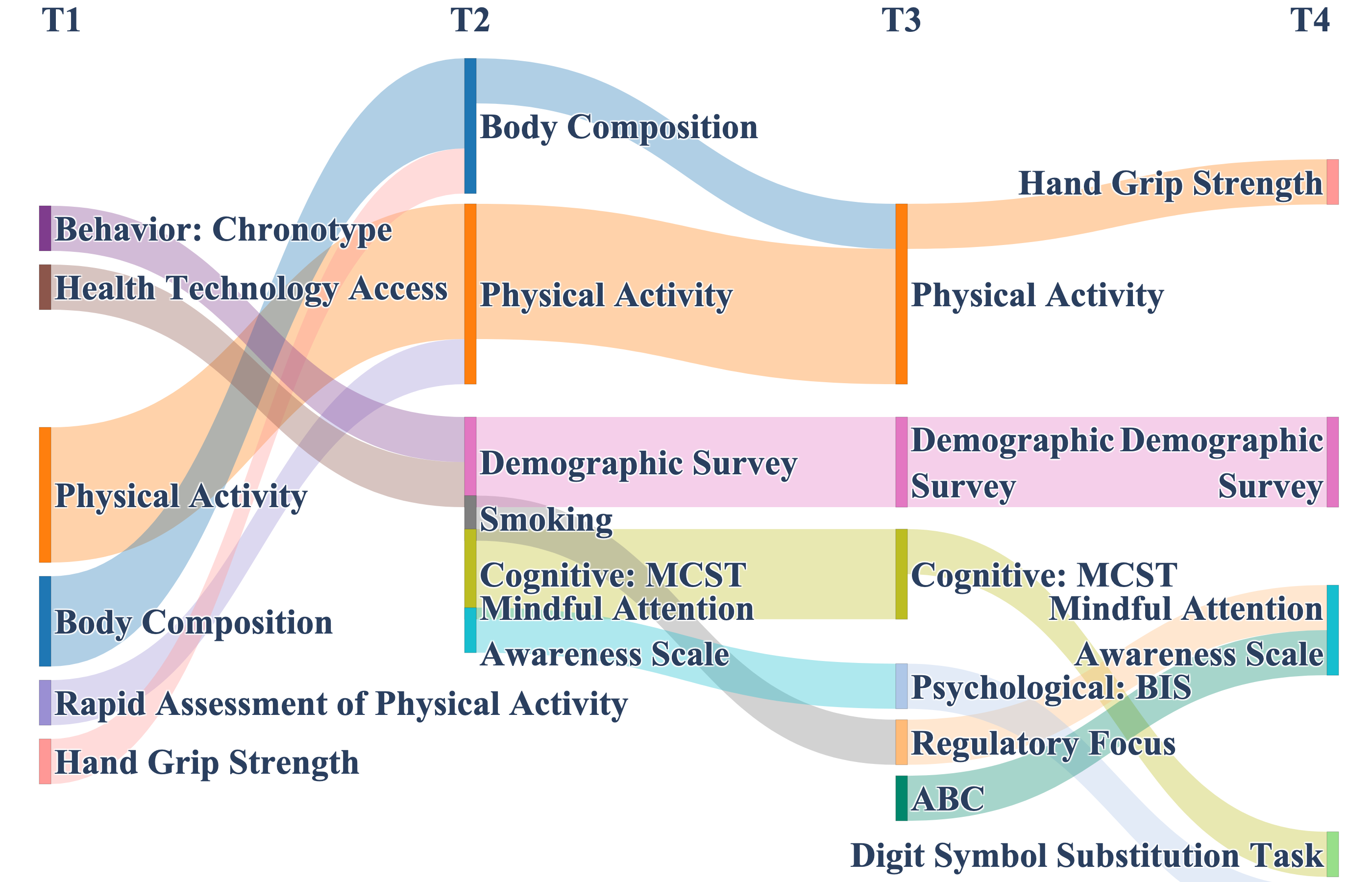} \label{subfig:peer}} 
 \hfill    
\subfigure[Control]{
\includegraphics[width=0.45\linewidth]{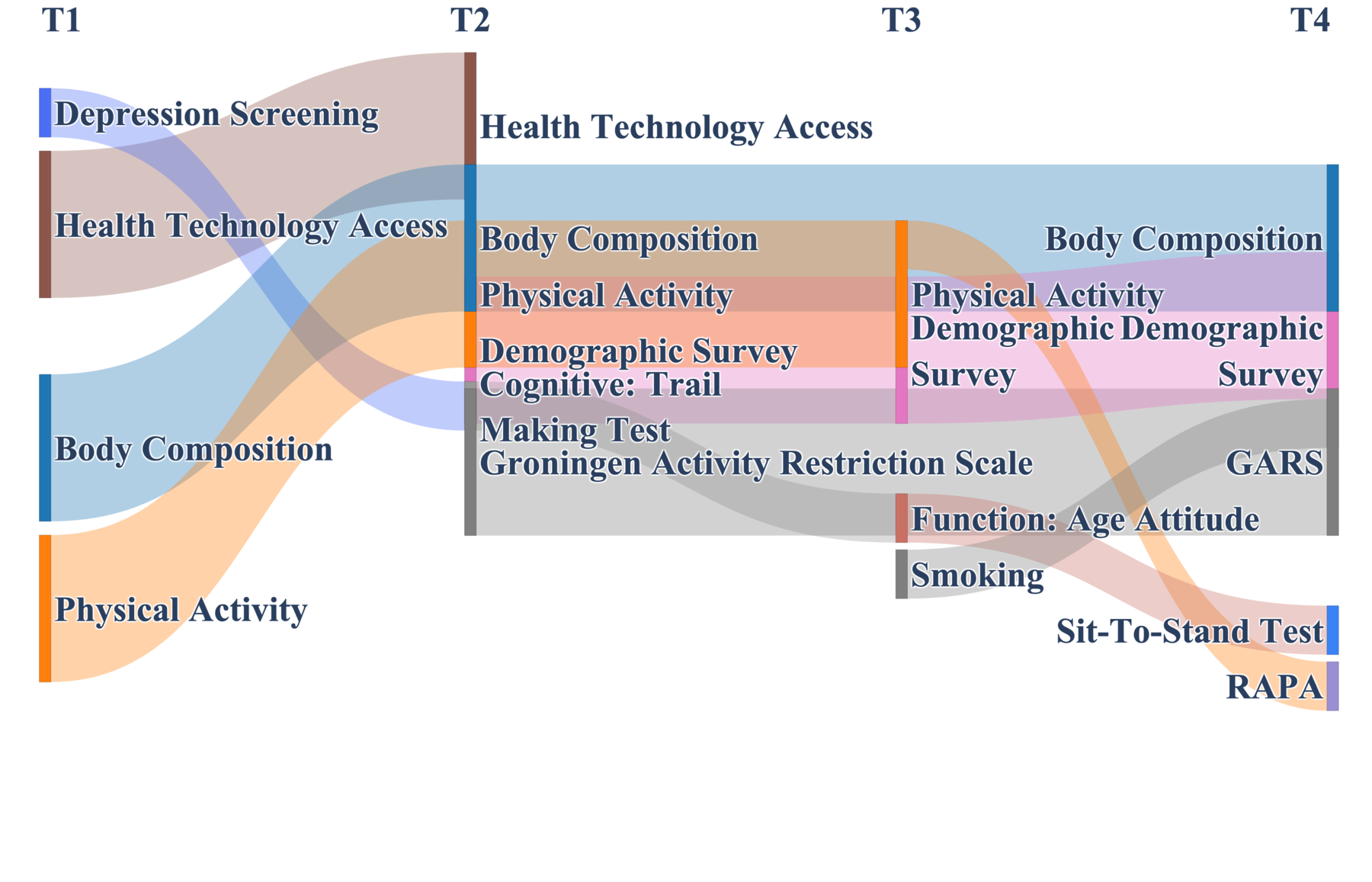}
    \label{subfig:control}}
\caption{Top 10 Fall Risk Factor Identification across 4 visits. (a) PEER Intervention, (b) Control. }
\label{fig:in-field_general}
\end{figure}

The top 10 most important fall risk factor groups and their evolving interconnections across four visits are visualized for both the PEER Intervention and Control in Fig.~\ref{fig:in-field_general}. In the PEER cluster (Fig.~\ref{fig:in-field_general}\subref{subfig:peer}), which received peer-led interventions, we observe greater temporal continuity in key domains such as cognitive function and physical activity. Fall risk factor groups like \textit{MCST}, \textit{Mindful Attention Awareness Scale}, and \textit{balance confidence} appear repeatedly across multiple visits, suggesting a consistent focus on cognitive engagement and behavioral monitoring throughout the intervention period. 

In contrast, the Control cluster shows more fragmented transitions, with higher variability in feature composition across visits in Fig.~\ref{fig:in-field_general}\subref{subfig:control}. Although core factors such as \textit{physical activity} and \textit{body composition} persist, other fall risk groups, particularly those related to cognition and behavior, tend to fluctuate more. This divergence highlights how the presence of a structured, peer-led program may help stabilize attention toward critical fall risk factors over time. 

Despite these differences, both clusters consistently highlight domains like \textit{physical activity} and \textit{body composition}, reinforcing their foundational importance in fall risk factors identification. However, the PEER cluster captures a broader range of psychological and cognitive factors, which may support more holistic and individualized monitoring. These patterns underscore the role of community-based interventions in fostering consistent and interpretable health data trajectories across longitudinal assessments. 

\subsection{Initial Reference Sets for Evaluation}\label{appendix:reference_sets}
To evaluate feature- and group-level selection performance, we use visit-specific initial reference sets defined from the graph- and node-embedding results. Let $\mathcal{R}_{\tau}^{\mathrm{feat}}$ and $\mathcal{R}_{\tau}^{\mathrm{grp}}$ denote the feature- and group-level reference sets at visit $\tau$, respectively. The reference sets are embedding-derived evaluation proxies constructed using the training partition only and independently of the evaluated methods. They provide a common reference for method comparison and should not be interpreted as exhaustive clinical ground truth.

\begin{table*}[htbp]
\centering
\caption{Initial feature- and group-level reference sets used for evaluation on the PEER study.}
\label{tab:initial_reference_sets}
\resizebox{\textwidth}{!}{
\begin{tabular}{lll}
\toprule
\textbf{Reference Group}
& \textbf{Reference Features}
& \textbf{Visits} \\
\midrule
Demographic Survey
& age, education, gender, health, living, number\_of\_falls, race,
sleeping\_pills, smoking
& T1--T4 \\

Fall Risk Screening: STEADI
& ste8
& T1--T4 \\

Depression Screening: PHQ-9
& phq
& T1--T4 \\

Anxiety Screening: GAI-SF
& gai\_sf\_score
& T1, T4 \\

Fear of Falling: Short FES-I
& fes
& T1--T4 \\

TUG, STS, and BTrackS
& bbs, tug
& T1--T4 \\

Short Physical Performance Battery
& balance, gait, speed\_gait, sppb, sts\_sppb\_3
& T1--T4 \\

Hand Grip Strength
& avg\_hgs\_kg\_both\_hands
& T1--T3 \\

Brief Aging Perceptions Questionnaire
& b\_apq\_chronic, b\_apq\_conseqeunce\_positive,
b\_apq\_control\_negative, b\_apq\_control\_positive, b\_apq\_score
& T2--T4 \\

Behavioral Inhibition/Activation System
& bas\_drive, bas\_fun, bas\_reward, bis\_score
& T4 \\

Regulatory Focus Questionnaire
& rfq\_prevention, rfq\_promotion
& T1, T4 \\

FRAIL Questionnaire
& fra
& T4 \\

Memory Impairment Screen
& mis
& T4 \\

RUDAS
& rudas\_score
& T4 \\
\bottomrule
\end{tabular}
}
\end{table*}

\subsection{Longitudinal Stability in the No-Fall Subgroup}
\label{appendix:no_fall_stability}

To evaluate the longitudinal stability of PAFIR, we compare the selected
feature groups across adjacent visits among participants with no recorded
falls during follow-up. For each participant, Jaccard similarity~\citep{hastie2009elements} is calculated as the number of feature groups selected at both visits divided by the total number of distinct feature groups selected across the two visits. Selection turnover~\citep{salop1976self,barrick2005reducing} is defined as one minus the Jaccard similarity, with lower values indicating fewer changes between visits.

For each random seed, Jaccard similarity and turnover are calculated for all eligible participants with valid selections at both visits and then averaged within each adjacent visit pair. The overall result is obtained by first averaging across the available adjacent visit pairs for each participant and then averaging across eligible participants. Table~\ref{tab:no_fall_stability} reports the mean and standard deviation of these results across 20 random seeds. Visit pairs with missing selections are excluded, and $N$ denotes the number of eligible no-fall participants included in each comparison.

\begin{table}[t]
\centering
\caption{Longitudinal stability of PAFIR group- and feature-level selections
among participants with no recorded falls. Results are reported as mean
$\pm$ standard deviation over 20 random seeds.}
\label{tab:no_fall_stability}
\resizebox{\linewidth}{!}{
\begin{tabular}{lcccccc}
\toprule
& & \multicolumn{2}{c}{\textbf{Group-Level Selection}}
& \multicolumn{2}{c}{\textbf{Feature-Level Selection}} \\
\cmidrule(lr){3-4}
\cmidrule(lr){5-6}
\textbf{Visit Pair}
& \textbf{$N$}
& \textbf{Jaccard $\uparrow$}
& \textbf{Turnover $\downarrow$}
& \textbf{Jaccard $\uparrow$}
& \textbf{Turnover $\downarrow$} \\
\midrule
T1--T2
& 253
& $0.744 \pm 0.007$
& $0.256 \pm 0.007$
& $0.095 \pm 0.007$
& $0.905 \pm 0.007$ \\

T2--T3
& 227
& $0.956 \pm 0.014$
& $0.044 \pm 0.014$
& $0.088 \pm 0.006$
& $0.912 \pm 0.006$ \\

T3--T4
& 158
& $0.969 \pm 0.015$
& $0.031 \pm 0.015$
& $0.064 \pm 0.007$
& $0.936 \pm 0.007$ \\
\midrule

Overall
& 253
& $0.890 \pm 0.009$
& $0.110 \pm 0.009$
& $0.083 \pm 0.004$
& $0.917 \pm 0.004$ \\
\bottomrule
\end{tabular}
}
\end{table}
Among the 341 participants in the analyzed cohort, 48 had at least one recorded fall and 293 had no recorded falls during follow-up. Among the no-fall participants, 253, 227, and 158 had valid PAFIR feature-group selections at both visits for T1--T2, T2--T3, and T3--T4, respectively. Overall, 253 participants contributed to at least one adjacent-visit comparison. As shown in Table~\ref{tab:no_fall_stability}, PAFIR exhibits distinct stability patterns at the group and feature levels. The overall group-level Jaccard similarity is $0.890 \pm 0.009$, with a corresponding turnover of $0.110 \pm 0.009$, indicating that the broader selected risk domains remain highly consistent across adjacent visits. In contrast, the overall feature-level Jaccard similarity is $0.083 \pm 0.004$, corresponding to a turnover of $0.917 \pm 0.004$. These results indicate that PAFIR maintains stable group-level selections while adaptively updating the specific features selected within those groups across visits.

\subsection{Feature- and Group-Level Benchmark and Ablation Results}\label{appendix:feature_group_results}
\begin{table*}[t]
\centering
\caption{Feature- and group-level fall-risk factor selection results on the PEER study. Precision, recall, and F1 score are computed relative to the predefined feature- and group-level reference sets.}
\label{tab:fall_risk_node_group}
\resizebox{\textwidth}{!}
{
\begin{tabular}{c|ccc|ccc}
\toprule
\textbf{Method}
& \multicolumn{3}{c|}{\textbf{Feature-Level Selection}}
& \multicolumn{3}{c}{\textbf{Group-Level Selection}} \\
\cmidrule(lr){2-4} \cmidrule(lr){5-7}
& Precision $\uparrow$
& Recall $\uparrow$
& F1 Score $\uparrow$
& Precision $\uparrow$
& Recall $\uparrow$
& F1 Score $\uparrow$ \\
\midrule
\textbf{PAFIR (Ours)}
& 0.887 $\pm$ 0.008
& \textbf{0.874 $\pm$ 0.007}
& 0.880 $\pm$ 0.007
& \textbf{0.781 $\pm$ 0.011}
& \textbf{0.793 $\pm$ 0.009}
& \textbf{0.787 $\pm$ 0.010}
\\
GraphGPS+HDSE~(\citeyear{luo2024enhancing})
& \textbf{0.891 $\pm$ 0.009}
& \textbf{0.874 $\pm$ 0.008}
& \textbf{0.882 $\pm$ 0.008}
& 0.736 $\pm$ 0.009
& 0.744 $\pm$ 0.010
& 0.740 $\pm$ 0.009
\\
TAR~(\citeyear{ying2024topology})
& 0.788 $\pm$ 0.010
& 0.795 $\pm$ 0.009
& 0.791 $\pm$ 0.010
& 0.767 $\pm$ 0.008
& 0.791 $\pm$ 0.008
& 0.779 $\pm$ 0.008
\\
SAT~(\citeyear{chen2022structure})
& 0.723 $\pm$ 0.009
& 0.731 $\pm$ 0.010
& 0.727 $\pm$ 0.009
& 0.677 $\pm$ 0.013
& 0.689 $\pm$ 0.011
& 0.683 $\pm$ 0.012
\\
PCA~(\citeyear{uddin2021pca})
& 0.643 $\pm$ 0.013
& 0.624 $\pm$ 0.015
& 0.638 $\pm$ 0.014
& 0.573 $\pm$ 0.013
& 0.566 $\pm$ 0.012
& 0.569 $\pm$ 0.011
\\

TTG~(\citeyear{khurana2018feature})
& 0.677 $\pm$ 0.008
& 0.664 $\pm$ 0.009
& 0.670 $\pm$ 0.008
& 0.643 $\pm$ 0.012
& 0.621 $\pm$ 0.012
& 0.611 $\pm$ 0.012
\\
\bottomrule
\end{tabular}
}
\vspace{-0.08in}
\end{table*}

Table~\ref{tab:fall_risk_node_group} reports feature- and group-level
selection performance on the PEER study. PAFIR achieves the highest
group-level precision, recall, and F1 score, indicating that the hierarchical selection policy effectively identifies clinically relevant feature groups. At the individual-feature level, GraphGPS+HDSE obtains slightly higher precision and F1 score, while PAFIR achieves comparable recall. The ablation results show that removing hierarchical graph encoding, temporal encoding, or the three-agent comparison-driven selection mechanism reduces feature-level performance. Removing the three-agent mechanism also decreases group-level F1 from 0.787 to 0.716, supporting the contribution of the candidate, reference, and operation agents to group-level feature selection.

\subsection{Computational Efficiency}
\label{appendix:computational_efficiency}

\begin{table}[t]
\centering
\caption{Computational cost of PAFIR on the PEER study.}
\label{tab:runtime}
\begin{tabular}{lc}
\toprule
\textbf{Measurement} & \textbf{Result} \\
\midrule
GPU & NVIDIA Quadro RTX 6000 \\
GPU memory capacity & 24 GB \\
CUDA version & 13.0 \\
Runtime per random seed & $\sim$101 s \\
Runtime for 20 seeds & $\sim$33.7 min \\
Peak allocated GPU memory & $\sim$240 MB \\
\bottomrule
\end{tabular}
\end{table}

\section{Preprocessing}\label{appendix:preprocess}
\subsection{Hierarchical State Representation}\label{appendix:encoding}
To effectively model both the structural health assessment and temporal dynamics in longitudinal health data, we propose a state representation $\mathbf{s}_{i,t}$ that integrates graph-based and time-series modalities via a cross-attention mechanism. This architecture captures health status evolution across visits through three key components that jointly encode structurally informed features, such as fear of falling, at each visit, and temporally aligned physical activity patterns, such as step counts and vector magnitudes.

\paragraph{Hierarchical Structural Encoding}
First, structurally informed features are derived by applying the hierarchical distance structure encoding (HDSE)~\citep{luo2024enhancing}, which reveals and encodes latent multi-level feature relationships among features and their association with the primary outcome, \textit{fall incidence}. At each visit $\tau$, each node represents a feature (e.g., \textit{BMI}, \textit{weight}, and \textit{height}), and each feature group (e.g., \textit{body composition}) forms a fully connected graph $\mathcal{G}_m = (\mathcal{V}_m, \mathcal{E}_m)$, where $\mathcal{V}_m$ is the set of features belonging to group $m$. We construct a multi-level hierarchy for each feature group by iteratively coarsening the graph to generate $K+1$ levels. At each level $k$, we compute the graph hierarchy distance (GHD)~\citep{luo2024enhancing} between nodes $v \in \mathcal{V}_m$ and $u \in \mathcal{V}_m$, denoted as,
\begin{align*}
    \text{GHD}^{k}(v,u)
    =
    \text{SPD}\!\left(
    \psi_{k-1} \circ \cdots \circ \psi_0(v),
    \psi_{k-1} \circ \cdots \circ \psi_0(u)
    \right),
\end{align*}
where $\psi_{k-1} \circ \cdots \circ \psi_0(v)$ is the mapping of node $v$ from level $0$ to level $k$ in the graph hierarchy, and $\text{SPD}(\cdot, \cdot)$ denotes shortest path distance. 

For each feature node $v$, we construct a structural distance tensor $\mathbf{D}_v \in \mathbb{R}^{(K+1) \times |\mathcal{V}_m|}$ by stacking its pairwise distances to all other nodes $u \in \mathcal{V}_m$ across all hierarchy levels. Each entry $\mathbf{D}_{v,u}$ captures the multi-level distances from node $v$ to node $u$,
\begin{equation}
\label{multi-structure-distance}
\mathbf{D}_{v,u}
=
\left[
\text{GHD}^{0}(v,u),
\text{GHD}^{1}(v,u),
\cdots,
\text{GHD}^{K}(v,u)
\right].
\end{equation}
This tensor captures the multi-level structural relationships of node $v$ within group $m$. To obtain a compact representation, we apply positional encoding to $\mathbf{D}_v$, yielding a dense embedding that encodes the hierarchical position of the feature.

\subsection{Temporal Evolving for Sequence and Time Series}
\label{appendix:timeencoding}
To capture the physical activity dynamics observed after each clinical visit, we adopt a temporal state encoder based on the iTransformer architecture~\citep{liu2023itransformer}. For each feature, such as step counts and VMs, we extract a 7-day time series collected starting from the visit day, and treat it as an input token. 

Let $\mathbf{X}_{i,:,b}^{(\tau)} \in \mathbb{R}^{L \times 1}$ denote the time series of feature $b$ during the 7-day period following visit $\tau$ for participant $i$, where $L$ is the sequence length (e.g., minute-level for 7 days). Each sequence is projected into a latent space via a shared embedding function,
\begin{equation}
\label{temporal-embed}
\mathbf{z}_{i,t,b}^{(0)}
=
\text{Embed}\!\left(
\mathbf{X}_{i,:,b}^{(\tau)}
\right)
\in
\mathbb{R}^{d},
\end{equation}
where $d$ means the embedding dimension, which controls the capacity of the latent representation. 
We then apply a stack of $M_{\mathrm{temp}}$ Transformer encoder blocks to capture temporal dependencies,
\begin{equation}
\mathbf{z}_{i,t,b}^{(\ell+1)}
=
\text{TrmBlock}^{(\ell)}
\left(
\mathbf{z}_{i,t,b}^{(\ell)}
\right),
\qquad
\ell=0,\ldots,M_{\mathrm{temp}}-1.
\end{equation}
The final encoder output $\mathbf{z}_{i,t,b}$ summarizes the temporal dynamics of feature $b$ and serves as the query for the cross-attention mechanism.

\paragraph{Cross-attention Mechanism}
In the cross-attention mechanism, we regard the temporal embeddings of physical activity patterns as queries
$\mathbf{Q}_{i,t} \in \mathbb{R}^{|\mathcal{B}| \times d'}$,
the structural embeddings of feature groups as keys
$\mathbf{K}_{i,t} \in \mathbb{R}^{M \times d'}$,
and the corresponding within-group structural embeddings as values
$\mathbf{V}_{i,t} \in \mathbb{R}^{M \times d'}$,
where $\mathcal{B}$ is the set of temporal features, $M$ is the number of feature groups, and $d'$ is the hidden dimension per attention head. To incorporate structural priors, we introduce a bias matrix
$\mathbf{B}_{i,t}^{\mathrm{struct}}
\in
\mathbb{R}^{|\mathcal{B}| \times M}$
to encourage each temporal query to attend more strongly to structurally relevant feature groups, where each entry encodes the relative GHD-based proximity between temporal feature $b$ and feature group $\mathcal{V}_m$. The attention is then computed as
\begin{equation}
\text{CrossAttn}
\left(
\mathbf{Q}_{i,t},
\mathbf{K}_{i,t},
\mathbf{V}_{i,t}
\right)
=
\text{softmax}
\left(
\frac{
\mathbf{Q}_{i,t}\mathbf{K}_{i,t}^{\top}
}{
\sqrt{d'}
}
+
\mathbf{B}_{i,t}^{\mathrm{struct}}
\right)
\mathbf{V}_{i,t}.
\end{equation}
This fusion mechanism enables the model to align temporal physical activity patterns with the longitudinal structure of clinical survey data, allowing for more informed and interpretable inter-feature association.

\subsection{Three-Agent RL Feature Selection Framework}

We leverage a three-agent decision framework~\citep{ying2024topology} that performs personalized and temporally adaptive feature selection under RL. At each clinical visit, the state representation $\mathbf{s}_{i,t}$ incorporates both temporal behavior dynamics from physical activity and structural information from clinical feature topology, and is used to guide the decision-making in a three-agent RL structure composed of a candidate feature group agent, a reference feature group agent, and an operation agent. In the RL framework, time step $t$ can correspond to a clinical visit $\tau$ or to the occurrence of a fall incident for participant $i$.
Let
$\mathcal{C}
=
\{\mathcal{V}_1,\mathcal{V}_2,\dots,\mathcal{V}_M\}$
denote the collection of clinically defined feature groups.

\paragraph{Group-Level Feature Selection} 
At each iteration $t$, the framework operates
\begin{itemize}[noitemsep,topsep=0pt,parsep=0pt,partopsep=0pt]
    \item \textit{Candidate Feature Group Agent}: Selects a feature group $C_{i,t}^{\mathrm{cand}}\in\mathcal{C}$ based on the current state $\mathbf{s}_{i,t}$.
    
    \item \textit{Reference Feature Group Agent}: Selects a comparison feature group $C_{i,t}^{\mathrm{ref}} \in \mathcal{C}\setminus \{C_{i,t}^{\mathrm{cand}}\}$ from the remaining feature groups.
    
    \item \textit{Operation Agent}: Selects an add, remove, or retain operation based on the relative incremental utility of $C_{i,t}^{\mathrm{cand}}$ and $C_{i,t}^{\mathrm{ref}}$ under the available outcome signal $y_{i,t}$. 
\end{itemize}

The resulting action $a_{i,t} =
(C_{i,t}^{\mathrm{cand}},C_{i,t}^{\mathrm{ref}},o_{i,t})$ governs the inclusion or exclusion of existing feature groups. The group with higher relevance to the outcome is retained in the selected feature set $F_{i,t}$, and the other is used as the reference in the next iteration.

\paragraph{Within-Group Feature Selection}
To further refine feature granularity, we extend the three-agent structure to operate within each selected group $\mathcal{V}_m \in F_{i,t}^{\mathrm{group}}$. For each group, a second round of selection is performed at the individual feature level. A localized state representation $\mathbf{s}_{i,t}^{(\mathcal{V}_m)}$ encodes the temporal and structural context of individual features in $\mathcal{V}_m$.

Within-group agents operate similarly.  The \underline{Candidate Feature Agent} selects a feature $x_{i,t}^{\mathrm{cand}}\in\mathcal{V}_m$ (e.g., \textit{BMI}), while the \underline{Reference Feature Agent} chooses a comparison feature $x_{i,t}^{\mathrm{ref}}\in\mathcal{V}_m$ (e.g., \textit{fat-free mass}) from the remaining features in the same group. Finally, the \underline{Operation Agent} decides whether to retain $x_{i,t}^{\mathrm{cand}}$.

The temporal-difference (TD) learning and reward mechanisms follow the same procedure as the group level. The Q-networks are randomly initialized, while the HDSE node-level importance scores are used to initialize the feature-level rewards.

\section{Benchmarking and Results}\label{appedix:benchmark}
In this section, we evaluate our proposed method on ten public datasets and demonstrate state-of-the-art performance. We investigate the research problems: {RQ1}: Can our method effectively reconstruct high-quality feature spaces to improve performance on downstream tasks in both graph and time series domains? {RQ2}: Does our unified architecture outperform state-of-the-art baselines in terms of both structural and temporal evaluation metrics across diverse datasets? 

\subsection{Baselines} We compare our method against state-of-the-art baselines. For graph-based benchmarks, we include HDSE~\citep{luo2024enhancing}, TAR~\citep{ying2024topology}, SAT~\citep{chen2022structure}, PCA~\citep{uddin2021pca}, and TTG~\citep{khurana2018feature}. For time series forecasting, we consider iTransformer~\citep{liu2023itransformer}, RLinear~\citep{li2023revisiting}, PatchTST~\citep{nie2022time}, TimesNet~\citep{wu2022timesnet}, and SCINet~\citep{liu2022scinet}.

\subsection{Benchmark Dataset Description}
We evaluate our method on six public benchmark datasets across different domains. For graph-based comparisons, we use two bioinformatics datasets: ENZYMES~\citep{schomburg2004brenda} and PROTEINS~\citep{dobson2003distinguishing}, and a small molecules dataset: AIDS~\citep{riesen2008iam}.
For time series evaluation, we use three datasets from the energy and transportation domains: Solar-Energy~\citep{lai2018modeling}, Traffic~\citep{wu2021autoformer}, Weather~\citep{wu2021autoformer}, and ETT~\citep{zhou2021informer} (details in Table~\ref{tab:data_detials}.).

\begin{table*}[ht]
\centering
\caption{Detailed Dataset Description. Dim denotes the variate number of each dataset. Dataset Size denotes the total number of time points in (Train, Validation, Test), respectively. Prediction Length denotes the future time points to be predicted in each dataset. Frequency denotes the sampling interval of timepoints. \textit{Top Side}: Time Series Public Datasets. \textit{Bottom Side}: Graph Public Datasets.}

\resizebox{\textwidth}{!}{
\begin{tabular}{c|c|c|c|c|c}
\toprule
\textbf{Time-Series Dataset} & \textbf{Dim} & \textbf{Prediction Length} & \textbf{Dataset Size}& \textbf{Frequency} & \textbf{Information} \\
\midrule
ETTh1, ETTh2 & 7 & \{96, 192, 336, 720\} & (8545, 2881, 2881) & Hourly & Electricity \\
ETTm1, ETTm2 & 7 & \{96, 192, 336, 720\} & (34465, 11521, 11521) & 15min & Electricity \\
Traffic &862 &  \{96, 192, 336, 720\} & (12185, 1757, 3509) & Hourly & Transportation\\
Solar-Energy &137 & \{96, 192, 336, 720\} & (36601, 5161, 10417) & Hourly & Energy\\
Weather & 21 & \{96, 192, 336, 720\} & (36792, 5271, 10540) & 10min & Weather\\
\midrule
\textbf{Graph Dataset} & \textbf{Graphs Counts} & \textbf{Nodes Counts} & \textbf{Graph Classes}& \textbf{ Node Classes} & \textbf{Labels} \\
\midrule
ENZYMES & 600 & 19580 & 6 & 3 & Yes\\
PROTEINS & 1113 & 43471 & 2 & 3 & Yes\\
AIDS & 2000 & 31385 & 2 & 38 & Yes\\
\bottomrule
\end{tabular}
}
\label{tab:data_detials}
\end{table*}

For the graph prediction, we follow the data processing and train-validate-test set split protocol used in TAR~\cite{ying2024topology}. For the time-series prediction, we follow the same data processing and train-validate-test set split protocol used in iTransformer~\cite{liu2023itransformer}, where the train, validation, and test datasets are strictly divided according to chronological order to make sure there are no data leakage issues.

\subsection{Environment Setup and Metrics}
For graph benchmarks, we follow the data-processing and split protocol of TAR~\citep{ying2024topology}. For time-series benchmarks, we use the predefined chronological train/validation/test splits adopted by iTransformer~\citep{liu2023itransformer}. To ensure robustness, each experiment is repeated 10 times with different random seeds, and the mean performance is reported. For graph-based tasks, we evaluate the quality of the transformed feature space using Precision, Recall, and F1 score. Higher values indicate better performance. For reinforcement feature space reconstruction, training was limited to 10 epochs, each with 10 exploration steps. All agents were implemented using a DQN with two ReLU-activated linear layers, optimized with Adam (learning rate 0.01), an experience replay memory size of 32, and a batch size of 8. For time series forecasting tasks, we report prediction accuracy using Mean Squared Error (MSE) and Mean Absolute Error (MAE). Lower values indicate better performance. For the forecasting settings, we fixed the lookback window length to 96 for the ETT, Solar-Energy, Traffic, and Weather datasets, while the prediction horizons were set to vary among \{96, 192, 336, 720\}.

\paragraph{Model Regularization and Generalization}
Given the high-dimensional feature space and sparse fall outcomes, we adopt multiple strategies to mitigate overfitting in the proposed PAFIR framework. First, strong regularization is applied throughout the model, including dropout (rate = 0.5) and residual connections to stabilize deep representations. Second, hierarchical and group-wise structural representations reduce the effective dimensionality by leveraging clinically meaningful feature groupings rather than treating all variables independently.

For temporal modeling, a shared encoder is used across longitudinal visits, preventing visit-specific overfitting and encouraging the learning of consistent temporal patterns. Finally, the adaptive policy is optimized using temporally aggregated reward signals rather than directly fitting individual fall events, reducing sensitivity to sparse and delayed outcome labels.

\subsection{Benchmarks Results}  
To address {RQ1}, we evaluate the performance of our proposed method, PAFIR, in comparison with several state-of-the-art baselines across six benchmark datasets covering both graph classification and time series forecasting tasks. On the PROTEINS, ENZYMES, and AIDS datasets, as shown in Table~\ref{tab:graph_benchmark}, PAFIR achieves the highest scores across all evaluation metrics, including Precision, Recall, and F1 Score. These results confirm the effectiveness of PAFIR in capturing hierarchical structure patterns within graph data, consistently outperforming baselines such as GraphGPS+HDSE, TAR, SAT, PCA, and TTG.

\begin{table*}[ht]
\centering
\caption{Overall Performance Comparison on Hierarchical Graph Datasets. PAFIR outperforms existing methods on both node and graph classification tasks across PROTEINS, ENZYMES, and AIDS datasets, following the TAR~\citep{ying2024topology} setting.}
\setlength{\tabcolsep}{0pt}
\resizebox{\textwidth}{!}{
\begin{tabular}{c|c|ccc|ccc}
\toprule
\midrule
\textbf{Dataset} & \textbf{Method} 
& \multicolumn{3}{c|}{\textbf{Node Classification}} 
& \multicolumn{3}{c}{\textbf{Graph Classification}} \\
\cmidrule(lr){3-5} \cmidrule(lr){6-8}
 &  & Precision & Recall & F1 Score & Precision & Recall & F1 Score \\
\midrule

\parbox[c][0pt][c]{1.1em}{%
  \centering
  \rule{0pt}{10\baselineskip}%
  \rotatebox{90}{\textbf{PROTEINS}}%
}
 & PAFIR & \textbf{0.921 $\pm$ 0.007} & 0.914 $\pm$ 0.007 & \textbf{0.918 $\pm$ 0.009} & \textbf{0.843 $\pm$ 0.006} & \textbf{0.822 $\pm$ 0.007} & \textbf{0.832 $\pm$ 0.006} \\
& GraphGPS+HDSE~(\citeyear{luo2024enhancing}) & 0.867 $\pm$ 0.055 & 0.785 $\pm$ 0.097 & 0.780 $\pm$ 0.180 & 0.830 $\pm$ 0.027 & 0.812 $\pm$ 0.055 & 0.821 $\pm$ 0.012 \\
& TAR~\citeyear{ying2024topology} & 0.916 $\pm$ 0.017 & \textbf{0.925 $\pm$ 0.017} & 0.916 $\pm$ 0.017 & 0.767 $\pm$ 0.006 & 0.766 $\pm$ 0.007 & 0.765 $\pm$ 0.006 \\
& SAT (\citeyear{chen2022structure}) & 0.779 $\pm$ 0.104 & 0.785 $\pm$ 0.097 & 0.780 $\pm$ 0.037 & 0.769 $\pm$ 0.037 & 0.643 $\pm$ 0.039 & 0.701 $\pm$ 0.104 \\
& PCA~\citeyear{uddin2021pca} & 0.781 $\pm$ 0.003 & 0.719 $\pm$ 0.003 & 0.738 $\pm$ 0.003 & 0.643 $\pm$ 0.001 & 0.648 $\pm$ 0.002 & 0.644 $\pm$ 0.001 \\
& TTG~(\citeyear{khurana2018feature}) & 0.847 $\pm$ 0.005 & 0.856 $\pm$ 0.005 & 0.847 $\pm$ 0.005 & 0.752 $\pm$ 0.004 & 0.751 $\pm$ 0.004 & 0.751 $\pm$ 0.004 \\

\midrule
\parbox[c][0pt][c]{1.1em}{%
  \centering
  \rule{0pt}{10\baselineskip}%
  \rotatebox{90}{\textbf{ENZYMES}}%
}
& PAFIR & \textbf{0.937 $\pm$ 0.004} & \textbf{0.940 $\pm$ 0.007} & \textbf{0.941 $\pm$ 0.006} & \textbf{0.866 $\pm$ 0.006} & \textbf{0.848 $\pm$ 0.007} & \textbf{0.857 $\pm$ 0.006} \\
& GraphGPS+HDSE~(\citeyear{luo2024enhancing}) & 0.816 $\pm$ 0.006 & 0.862 $\pm$ 0.277 & 0.885 $\pm$ 0.121 & 0.842 $\pm$ 0.003 & 0.819 $\pm$ 0.004 & 0.831 $\pm$ 0.106 \\
& TAR~\citeyear{ying2024topology} & 0.936 $\pm$ 0.006 & 0.934 $\pm$ 0.005 & 0.937 $\pm$ 0.006 & 0.324 $\pm$ 0.053 & 0.358 $\pm$ 0.029 & 0.325 $\pm$ 0.029 \\
& SAT~(\citeyear{chen2022structure}) & 0.744 $\pm$ 0.007 & 0.782 $\pm$ 0.006 & 0.753 $\pm$ 0.004 & 0.694 $\pm$ 0.008 & 0.833 $\pm$ 0.037 & 0.757 $\pm$ 0.104 \\
& PCA~(\citeyear{uddin2021pca}) & 0.753 $\pm$ 0.000 & 0.768 $\pm$ 0.000 & 0.756 $\pm$ 0.000 & 0.239 $\pm$ 0.000 & 0.292 $\pm$ 0.000 & 0.260 $\pm$ 0.000 \\
& TTG~(\citeyear{khurana2018feature}) & 0.919 $\pm$ 0.003 & 0.928 $\pm$ 0.002 & 0.920 $\pm$ 0.003 & 0.257 $\pm$ 0.042 & 0.308 $\pm$ 0.015 & 0.261 $\pm$ 0.026 \\
\midrule
\parbox[c][0pt][c]{1.1em}{%
  \centering
  \rule{0pt}{8\baselineskip}%
  \rotatebox{90}{\textbf{AIDS}}%
}
& PAFIR & {0.987 $\pm$ 0.003} & \textbf{{0.991 $\pm$ 0.007}} & \textbf{0.989 $\pm$ 0.002} & \textbf{0.986 $\pm$ 0.001} & \textbf{0.985 $\pm$ 0.006} & \textbf{0.986 $\pm$ 0.001} \\
& GraphGPS+HDSE~(\citeyear{luo2024enhancing}) & 0.982 $\pm$ 0.002 & 0.989 $\pm$ 0.097 & 0.985 $\pm$ 0.021 & 0.984 $\pm$ 0.006 & \textbf{0.985 $\pm$ 0.004} & 0.984 $\pm$ 0.096 \\
& TAR~\citeyear{ying2024topology} & \textbf{0.988 $\pm$ 0.002} & \textbf{0.991 $\pm$ 0.001} & 0.988 $\pm$ 0.002 & 0.984 $\pm$ 0.002 & 0.984 $\pm$ 0.002 & 0.984 $\pm$ 0.002 \\
& SAT~(\citeyear{chen2022structure}) & 0.944 $\pm$ 0.007 & 0.982 $\pm$ 0.006 & 0.953 $\pm$ 0.004 & 0.946 $\pm$ 0.005 & 0.933 $\pm$ 0.017 & 0.957 $\pm$ 0.094 \\
& PCA~(\citeyear{uddin2021pca}) & 0.381 $\pm$ 0.000 & 0.615 $\pm$ 0.000 & 0.471 $\pm$ 0.000 & 0.899 $\pm$ 0.000 & 0.899 $\pm$ 0.000 & 0.893 $\pm$ 0.000 \\
& TTG~(\citeyear{khurana2018feature}) & 0.925 $\pm$ 0.023 & 0.947 $\pm$ 0.016 & 0.933 $\pm$ 0.020 & 0.896 $\pm$ 0.000 & 0.899 $\pm$ 0.000 & 0.893 $\pm$ 0.000 \\
\midrule
\bottomrule
\end{tabular}
}
\label{tab:graph_benchmark}
\end{table*}

For time series forecasting, we conduct experiments on Solar-Energy, Traffic, ETT (including ETTh1, ETTh2, ETTm1, and ETTm2), and Weather datasets with a fixed input sequence length of 96 and prediction lengths varying in \{96, 192, 336, 720\}. As summarized in Table~\ref{tab:time_series_benchmark}, PAFIR achieves the lowest MSE and MAE in most cases across all datasets, outperforming recent baselines including iTransformer, RLinear, PatchTST, TimesNet, and SCINet. These results demonstrate the model's strong ability to align temporal resolution with evolving feature dynamics. Overall, the consistent superiority of PAFIR in both graph and time series domains highlights its robustness and adaptability for complex learning tasks involving heterogeneous and temporal data.
\begin{table*}[ht]
\centering
\setlength{\tabcolsep}{1pt} %
\caption{Overall Performance for Time Series Forecasting. We compare extensive competitive methods following the setting of iTransformer~\citep{liu2023itransformer}. The input sequence length is set to 96 for all baselines.}
{\footnotesize
\newcommand{\DS}[2]{
\parbox[c][0pt][c]{1.1em}{%
  \centering
  \rule{0pt}{#1\baselineskip}%
  \rotatebox{90}{\textbf{#2}}%
}}
\resizebox{\textwidth}{!}{
\begin{tabular}{c|c|cc|cc|cc|cc|cc|cc}
\toprule
\textbf{Dataset} & \textbf{Seq.} 
& \multicolumn{2}{c|}{\textbf{PAFIR (Ours)}} 
& \multicolumn{2}{c|}{\textbf{iTransformer}(\citeyear{liu2023itransformer})} 
& \multicolumn{2}{c|}{\textbf{RLinear}(\citeyear{li2023revisiting})} 
& \multicolumn{2}{c|}{\textbf{PatchTST}(\citeyear{nie2022time})} 
& \multicolumn{2}{c|}{\textbf{TimesNet}(\citeyear{wu2022timesnet})} 
& \multicolumn{2}{c}{\textbf{SCINet}(\citeyear{liu2022scinet})} \\
\cmidrule(lr){3-4} \cmidrule(lr){5-6} \cmidrule(lr){7-8} \cmidrule(lr){9-10} \cmidrule(lr){11-12} \cmidrule(lr){13-14}
 &  & MSE & MAE & MSE & MAE & MSE & MAE & MSE & MAE & MSE & MAE & MSE & MAE \\
\midrule

\parbox[c][0pt][c]{1.3em}{%
  \centering
  \rule{0pt}{6\baselineskip}%
  \rotatebox{90}{%
    \shortstack[c]{\textbf{Solar}\\\textbf{Energy}}%
  }%
}
& 96  & 0.213 & $\mathbf{0.236}$ & $\mathbf{0.203}$ & 0.237 & 0.322 & 0.339 & 0.271 & 0.307 & 0.250 & 0.292 & 0.237 & 0.344 \\
& 192 & $\mathbf{0.226}$ & $\mathbf{0.244}$ & 0.233 & 0.261 & 0.359 & 0.356 & 0.267 & 0.310 & 0.296 & 0.416 & 0.281 & 0.383 \\
& 336 & $\mathbf{0.243}$ & $\mathbf{0.258}$ & 0.248 & 0.273 & 0.397 & 0.369 & 0.290 & 0.315 & 0.319 & 0.432 & 0.302 & 0.400 \\
& 720 & $\mathbf{0.246}$ & $\mathbf{0.268}$ & 0.250 & 0.276 & 0.397 & 0.356 & 0.289 & 0.317 & 0.338 & 0.425 & 0.310 & 0.400 \\
\midrule

\DS{5}{Traffic}
& 96  & $\mathbf{0.412}$ & $\mathbf{0.272}$ & 0.417 & 0.276 & 0.649 & 0.398 & 0.462 & 0.304 & 0.593 & 0.321 & 0.804 & 0.509 \\
& 192 & $\mathbf{0.424}$ & $\mathbf{0.269}$ & 0.428 & 0.282 & 0.601 & 0.366 & 0.466 & 0.296 & 0.617 & 0.336 & 0.789 & 0.505 \\
& 336 & $\mathbf{0.428}$ & $\mathbf{0.277}$ & 0.433 & 0.283 & 0.609 & 0.369 & 0.482 & 0.304 & 0.629 & 0.335 & 0.800 & 0.508 \\
& 720 & $\mathbf{0.451}$ & $\mathbf{0.296}$ & 0.467 & 0.301 & 0.647 & 0.387 & 0.514 & 0.322 & 0.645 & 0.351 & 0.841 & 0.523 \\
\midrule

\DS{6}{ETTh1}
& 96  & $\mathbf{0.381}$ & $\mathbf{0.398}$ & 0.383 & 0.405 & 0.386 & 0.400 & 0.414 & 0.419 & 0.384 & 0.402 & 0.654 & 0.599 \\
& 192 & $\mathbf{0.434}$ & $\mathbf{0.422}$ & 0.441 & 0.436 & 0.437 & 0.424 & 0.460 & 0.445 & 0.436 & 0.429 & 0.719 & 0.631 \\
& 336 & $\mathbf{0.433}$ & 0.450 & 0.487 & 0.458 & 0.479 & $\mathbf{0.446}$ & 0.501 & 0.466 & 0.491 & 0.469 & 0.778 & 0.659 \\
& 720 & 0.498 & 0.487 & 0.503 & 0.491 & $\mathbf{0.481}$ & $\mathbf{0.470}$ & 0.500 & 0.488 & 0.521 & 0.500 & 0.836 & 0.697 \\
\midrule

\DS{6}{ETTh2}
& 96  & $\mathbf{0.288}$ & 0.340 & 0.297 & 0.349 & \textbf{0.288} & \textbf{0.338} & 0.302 & 0.348 & 0.340 & 0.374 & 0.707 & 0.621 \\
& 192 & ${0.377}$ & ${0.398}$ & 0.380 & 0.400 & \textbf{0.374} & \textbf{0.390} & 0.388 & 0.400 & 0.402 & 0.414 & 0.860 & 0.689 \\
& 336 & 0.422 & 0.429 & 0.428 & 0.432 & \textbf{0.415} & $\mathbf{0.426}$ & 0.426 & 0.433 & 0.452 & 0.452 & 1.000 & 0.744 \\
& 720 & \textbf{0.419} & \textbf{0.438} & 0.427 & 0.445 & 0.420 & 0.440 & 0.431 & 0.446 & 0.462 & 0.468 & 1.249 & 0.838 \\
\midrule

\DS{6}{ETTm1}
& 96  & 0.330 & $\mathbf{0.366}$ & 0.334 & 0.368 & 0.355 & 0.376 & \textbf{0.329} & 0.367 & 0.338 & 0.375 & 0.418 & 0.438 \\
& 192 & 0.370 & $\mathbf{0.385}$ & 0.377 & 0.391 & 0.391 & 0.392 & \textbf{0.367} & \textbf{0.385} & 0.374 & 0.387 & 0.439 & 0.450 \\
& 336 & 0.408 & 0.413 & 0.426 & 0.420 & 0.424 & 0.415 & \textbf{0.399} & \textbf{0.410} & 0.410 & 0.411 & 0.490 & 0.485 \\
& 720 & 0.484 & 0.448 & 0.491 & 0.459 & 0.487 & 0.450 & \textbf{0.454} & \textbf{0.439} & 0.478 & 0.450 & 0.595 & 0.550 \\
\midrule

\DS{6}{ETTm2}
& 96  & $\mathbf{0.173}$ & $\mathbf{0.259}$ & 0.180 & 0.264 & 0.182 & 0.265 & 0.175 & \textbf{0.259} & 0.187 & 0.267 & 0.286 & 0.377 \\
& 192 & $\mathbf{0.241}$ & 0.305 & 0.250 & 0.309 & 0.246 & 0.304 & \textbf{0.241} & \textbf{0.302} & 0.249 & 0.309 & 0.399 & 0.445 \\
& 336 & $\mathbf{0.304}$ & {0.343} & 0.311 & 0.348 & 0.307 & \textbf{0.342} & 0.305 & 0.343 & 0.321 & 0.351 & 0.369 & \textbf{0.342} \\
& 720 & 0.407 & 0.403 & 0.412 & 0.407 & ${0.407}$ & $\mathbf{0.398}$ & \textbf{0.402} & 0.400 & 0.408 & 0.403 & 0.960 & 0.735 \\
\midrule

\DS{6}{Weather}
& 96  & $\mathbf{0.170}$ & $\mathbf{0.211}$ & 0.174 & 0.214 & 0.192 & 0.232 & 0.177 & 0.218 & 0.172 & 0.220 & 0.221 & 0.306 \\
& 192 & $\mathbf{0.206}$ & $\mathbf{0.244}$ & 0.221 & 0.254 & 0.240 & 0.271 & 0.225 & 0.259 & 0.219 & 0.261 & 0.261 & 0.340 \\
& 336 & $\mathbf{0.277}$ & \textbf{0.295} & 0.278 & 0.296 & 0.292 & 0.307 & 0.278 & 0.297 & 0.280 & 0.306 & 0.309 & 0.378 \\
& 720 & \textbf{0.351} & \textbf{0.340} & 0.358 & 0.349 & 0.364 & 0.353 & 0.354 & 0.348 & 0.365 & 0.359 & 0.377 & 0.427 \\
\bottomrule
\end{tabular}
}
\label{tab:time_series_benchmark}
}
\end{table*}

\section{Ablation Study} \label{appendix:ablation}
To evaluate the contribution of each key component in our unified architecture and address {RQ2}, we perform a comprehensive ablation study across both graph-based and time series forecasting tasks. Specifically, we examine the individual impact of (1) the hierarchical graph encoder, (2) the time series encoder, and (3) the multi-agent policy mechanism. Each component is systematically removed or replaced, and the resulting performance is compared against the full model. Experiments are conducted on representative benchmarks, including node classification, graph classification, and multivariate time series forecasting, to assess the effectiveness and generalizability of each architectural element. 

\subsection{Impact of Hierarchical Graph Encoding} 
To evaluate the effectiveness of the hierarchical graph encoder in capturing structural dependencies, we perform an ablation by removing this component from the PAFIR architecture. As shown in Table~\ref{tab:state_ablation}, the performance drops significantly across both node and graph classification tasks on the PROTEINS, ENZYMES, and AIDS datasets.

\begin{table*}[ht]
\centering
\setlength{\tabcolsep}{1pt}     
\caption{Ablation Experiment for Hierarchical Graph Encoding.}
{\footnotesize

\newcommand{\DS}[2]{
\parbox[c][0pt][c]{1.2em}{%
  \centering
  \rule{0pt}{#1\baselineskip}%
  \rotatebox{90}{\textbf{#2}}%
}}
\resizebox{\textwidth}{!}{
\begin{tabular}{c|c|ccc|ccc}
\toprule
\midrule
\textbf{Dataset} & \textbf{Model Variant}
& \multicolumn{3}{c|}{\textbf{Node Classification}}
& \multicolumn{3}{c}{\textbf{Graph Classification}} \\
\cmidrule(lr){3-5} \cmidrule(lr){6-8}
 &  & Precision & Recall & F1 Score & Precision & Recall & F1 Score \\
\midrule

\textbf{PROTEINS}
& PAFIR (Full)
& $\mathbf{0.921 \pm 0.007}$ & $\mathbf{0.914 \pm 0.007}$ & $\mathbf{0.918 \pm 0.009}$
& $\mathbf{0.843 \pm 0.006}$ & $\mathbf{0.822 \pm 0.007}$ & $\mathbf{0.832 \pm 0.006}$ \\
& \makecell[l]{w/o Hierarchical \\Graph Encoding}
& 0.678 $\pm$ 0.005 & 0.673 $\pm$ 0.013 & 0.675 $\pm$ 0.020
& 0.765 $\pm$ 0.033 & 0.757 $\pm$ 0.005 & 0.766 $\pm$ 0.003 \\
\midrule

\textbf{ENZYMES}
& PAFIR (Full)
& $\mathbf{0.937 \pm 0.004}$ & $\mathbf{0.940 \pm 0.007}$ & $\mathbf{0.941 \pm 0.006}$
& $\mathbf{0.866 \pm 0.006}$ & $\mathbf{0.848 \pm 0.007}$ & $\mathbf{0.857 \pm 0.006}$ \\
& \makecell[l]{w/o Hierarchical \\Graph Encoding}
& 0.926 $\pm$ 0.006 & 0.923 $\pm$ 0.005 & 0.927 $\pm$ 0.006
& 0.340 $\pm$ 0.009 & 0.344 $\pm$ 0.007 & 0.348 $\pm$ 0.001 \\
\midrule

\textbf{AIDS}
& PAFIR (Full)
& $\mathbf{0.987 \pm 0.003}$ & $\mathbf{0.991 \pm 0.007}$ & $\mathbf{0.989 \pm 0.002}$
& $\mathbf{0.986 \pm 0.001}$ & $\mathbf{0.985 \pm 0.006}$ & $\mathbf{0.986 \pm 0.001}$ \\
& \makecell[l]{w/o Hierarchical \\Graph Encoding}
& 0.896 $\pm$ 0.004 & 0.863 $\pm$ 0.006 & 0.884 $\pm$ 0.001
& 0.808 $\pm$ 0.006 & 0.820 $\pm$ 0.003 & 0.814 $\pm$ 0.009 \\
\midrule
\bottomrule
\end{tabular}
}
\label{tab:state_ablation}
}
\end{table*}

On PROTEINS, the F1 score for node classification drops from 0.918 to 0.675, while graph classification performance also declines from 0.832 to 0.766. The impact is even more pronounced on the ENZYMES dataset, where graph classification F1 decreases drastically from 0.857 to 0.348. On the AIDS dataset, although the overall performance remains high, the consistent drop in F1 scores for both node and graph classification after removing hierarchical encoding further confirms its role in stabilizing and enhancing structural representation learning. This suggests that hierarchical structural information plays a crucial role in enabling PAFIR to identify task-relevant topological and semantic patterns.

The results confirm that incorporating multi-level structural representations allows the model to generalize better across diverse graph distributions and enhances its capacity to model complex relationships between features and labels.

\begin{figure}[ht]
\centering
\subfigure[Node Classification]{%
    \includegraphics[width=0.48\linewidth]{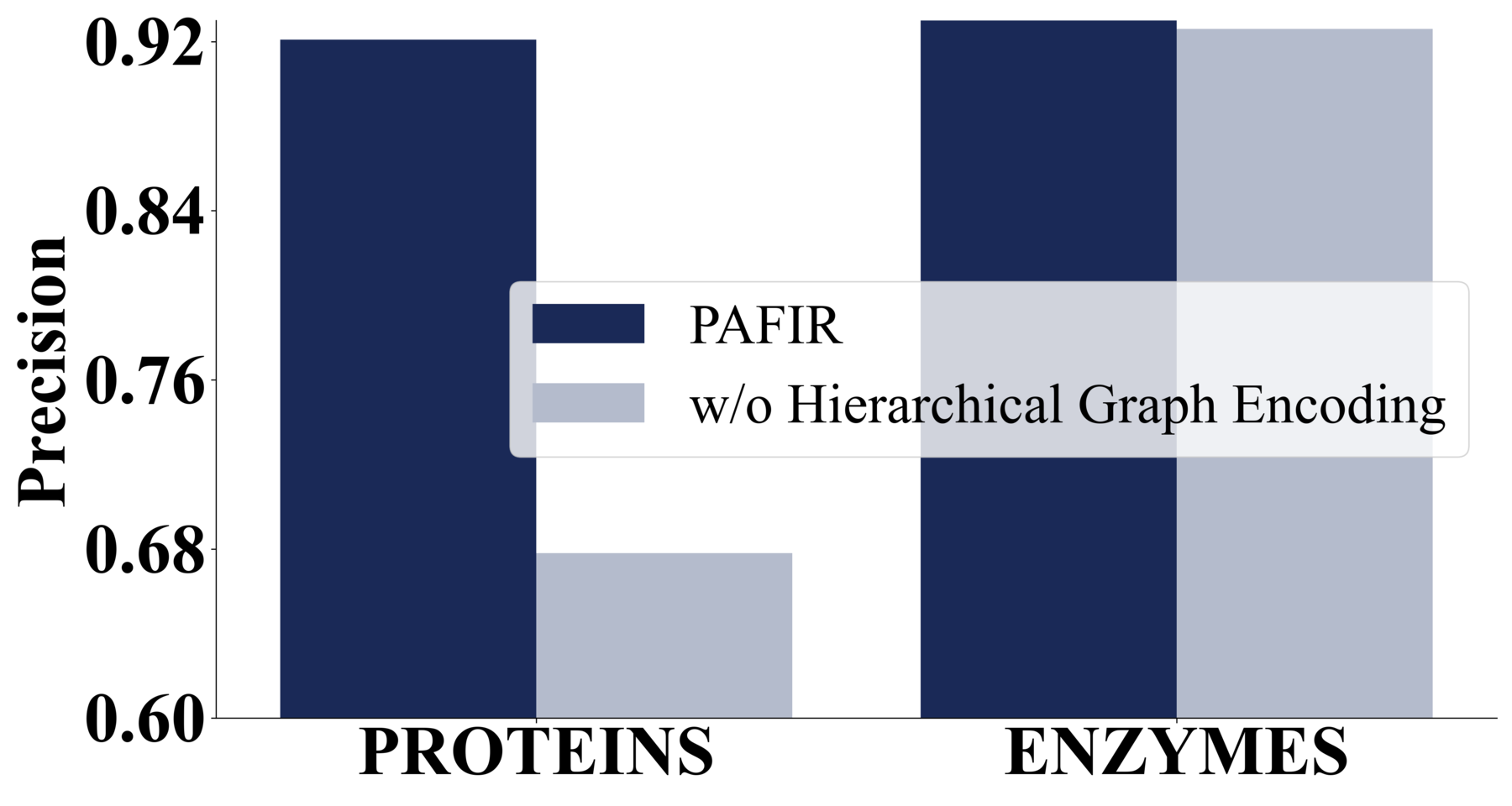}
}
\hfill
\subfigure[Graph Classification]{%
    \includegraphics[width=0.48\linewidth]{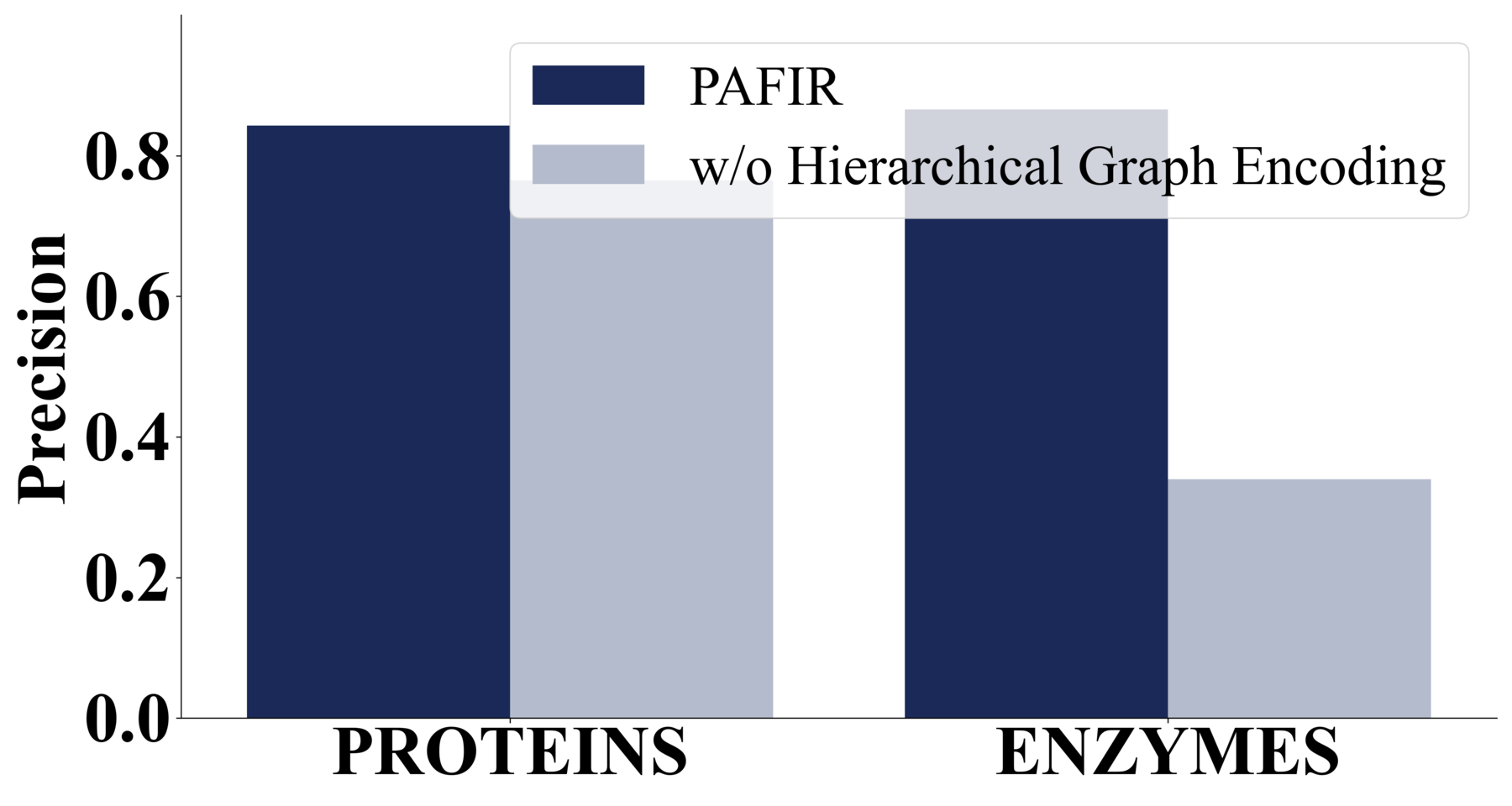}
}
\caption{Ablation experiment on the graph domain. We report Precision under two tasks: node classification and graph classification.}
    \vspace{-0.1in}
\label{fig:state_ablation}
\end{figure}

\subsection{Impact of Temporal Encoding}
To assess the contribution of the time series encoding module in capturing temporal dynamics, we remove it from the PAFIR architecture and compare performance across four forecasting benchmarks: Solar-Energy, Traffic, ETT, and Weather. As shown in Table~\ref{tab:time_series_ablation}, removing the temporal encoder results in substantial performance degradation across all datasets.

\begin{table*}[ht]
\centering
\caption{Ablation Experiment for Time Series Encoding. The input sequence length is fixed to 96 for all methods.}
\setlength{\tabcolsep}{0pt}
\resizebox{\textwidth}{!}{
\begin{tabular}{c|cc|cc|cc|cc|cc|cc|cc}
\toprule
\midrule
\textbf{Model Variant} 
& \multicolumn{2}{c|}{\textbf{Solar-Energy}} 
& \multicolumn{2}{c|}{\textbf{Traffic}}
& \multicolumn{2}{c|}{\textbf{ETTh1}} 
& \multicolumn{2}{c|}{\textbf{ETTh2}}
& \multicolumn{2}{c|}{\textbf{ETTm1}}
& \multicolumn{2}{c|}{\textbf{ETTm2}}
& \multicolumn{2}{c}{\textbf{Weather}} \\
\cmidrule(lr){2-3} \cmidrule(lr){4-5} \cmidrule(lr){6-7} \cmidrule(lr){8-9} \cmidrule(lr){10-11} \cmidrule(lr){12-13} \cmidrule(lr){14-15}
&MSE & MAE &  MSE & MAE &  MSE & MAE &  MSE & MAE &  MSE & MAE &  MSE & MAE &  MSE & MAE\\
\midrule
PAFIR (Full) 
 & $\mathbf{0.213}$ & $\mathbf{0.236}$ & $\mathbf{0.412}$ & $\mathbf{0.272}$ & $\mathbf{0.381}$ & $\mathbf{0.398}$ & \textbf{0.288} & \textbf{0.340} & \textbf{0.330} & \textbf{0.366} & \textbf{0.173} & \textbf{0.259} & \textbf{0.170} & \textbf{0.211}\\
w/o Time Series Encoding 
& 0.716  & 0.754 & 0.788  & 0.761  & 0.725  & 0.739 & 0.733 & 0.748 & 0.656 & 0.671 & 0.459 & 0.470 & 0.433 & 0.451\\
\midrule
\bottomrule
\end{tabular}
}
\label{tab:time_series_ablation}
\end{table*}

For instance, on the Solar-Energy dataset with a sequence length of 96, removing time series encoding causes the MSE to increase from 0.213 to 0.716 and the MAE from 0.236 to 0.754. Consistent and substantial performance degradations are also observed across Traffic, ETTh1, ETTh2, ETTm1, ETTm2, and Weather datasets, where both MSE and MAE increase markedly, in several cases by nearly two to three times. These results demonstrate that the temporal encoding module plays a crucial role in capturing sequential dependencies and aligning feature evolution with predictive accuracy across diverse time-series forecasting tasks.

\begin{figure}[ht]
    \centering
    \includegraphics[width=0.6\linewidth]{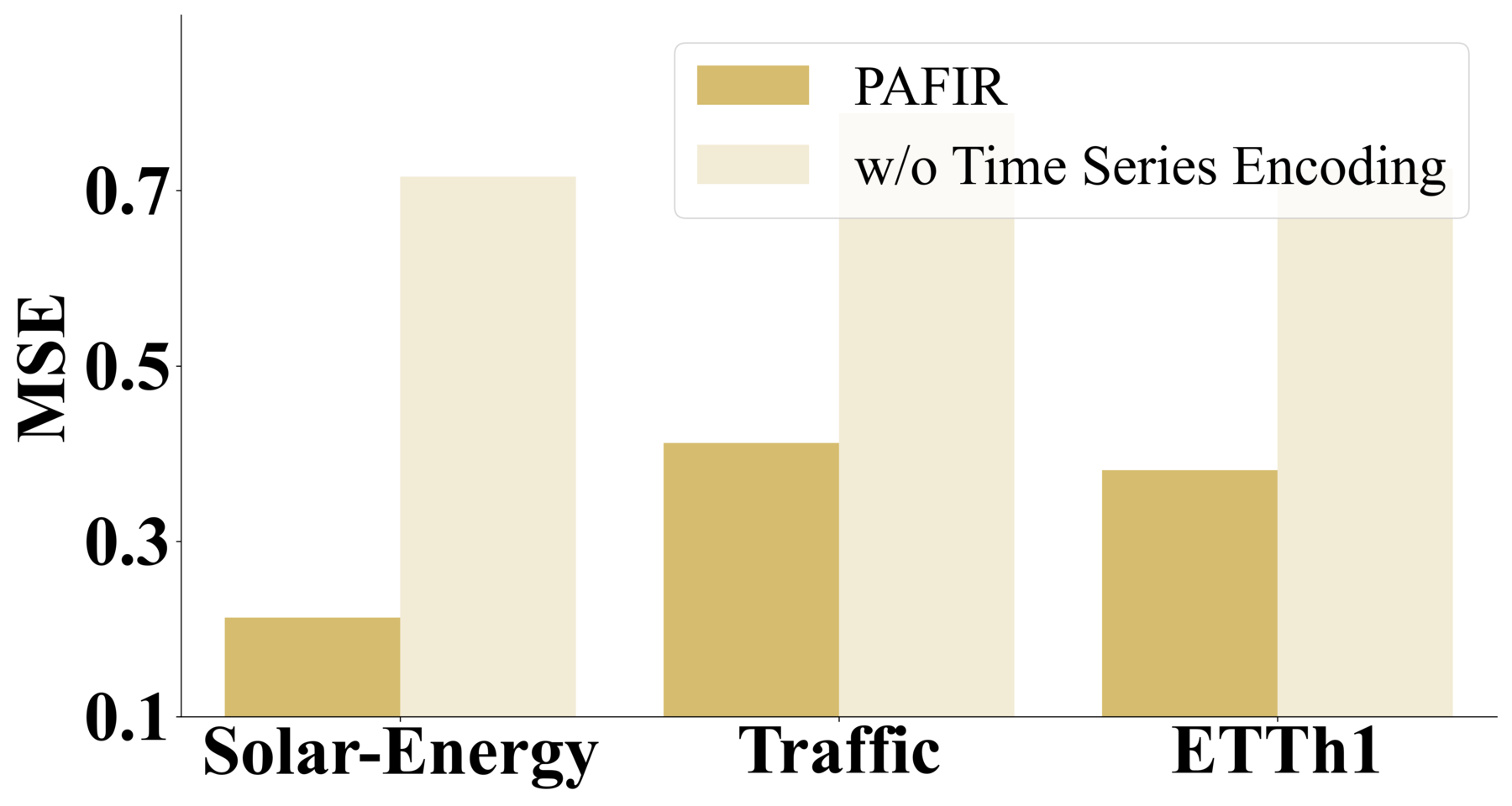}
    \caption{Ablation Experiment on Time Series Domain. We report MSE under Three Datasets.}
    \vspace{-0.1in}
    \label{fig:time_series_ablation}
\end{figure}

\subsection{Impact of Multi-Agent Policy}
To investigate the role of the multi-agent policy in PAFIR, we conduct an ablation by removing this component and evaluating performance on both node and graph classification tasks. As presented in Table~\ref{tab:agent_ablation}, removing the multi-agent policy leads to a notable drop in graph classification performance, while the impact on node classification is relatively modest.

\begin{table*}[ht]
\centering
\caption{Ablation Experiment for Multi-Agent Module in RL Framework.}
\setlength{\tabcolsep}{1pt}
\resizebox{\textwidth}{!}{
\begin{tabular}{c|c|ccc|ccc}
\toprule
\midrule
\textbf{Dataset} & \textbf{Model Variant} 
& \multicolumn{3}{c|}{\textbf{Node Classification}} 
& \multicolumn{3}{c}{\textbf{Graph Classification}} \\
\cmidrule(lr){3-5} \cmidrule(lr){6-8}
 &  & Precision & Recall & F1 Score & Precision & Recall & F1 Score \\
\midrule
\textbf{PROTEINS}
& PAFIR (Full) & $\mathbf{0.921 \pm 0.007}$ & $\mathbf{0.914 \pm 0.007}$ & $\mathbf{0.918 \pm 0.009}$ & $\mathbf{0.843 \pm 0.006}$ & $\mathbf{0.822 \pm 0.007}$ & $\mathbf{0.832 \pm 0.006}$ \\
& w/o Multi-Agent Policy & 0.916 $\pm$ 0.003 & ${0.913 \pm 0.002}$ & 0.917 $\pm$ 0.002 & 0.761 $\pm$ 0.010 & 0.760 $\pm$ 0.007 & 0.759 $\pm$ 0.007 \\
\midrule
\textbf{ENZYMES}
& PAFIR (Full) & $\mathbf{0.937 \pm 0.004}$ & $\mathbf{0.940 \pm 0.007}$ & $\mathbf{0.941 \pm 0.006}$ & $\mathbf{0.866 \pm 0.006}$ & $\mathbf{0.848 \pm 0.007}$ & $\mathbf{0.857 \pm 0.006}$ \\
& w/o Multi-Agent Policy & 0.926 $\pm$ 0.003 & 0.932 $\pm$ 0.002 & 0.927 $\pm$ 0.003 & 0.339 $\pm$ 0.040 & 0.348 $\pm$ 0.002 & 0.306 $\pm$ 0.004 \\
\midrule
\textbf{AIDS}
& PAFIR (Full) & $\mathbf{0.987 \pm 0.003}$ & $\mathbf{0.991 \pm 0.007}$ & $\mathbf{0.989 \pm 0.002}$ & $\mathbf{0.986 \pm 0.001}$ & $\mathbf{0.985 \pm 0.006}$ & $\mathbf{0.986 \pm 0.001}$ \\
& w/o Multi-Agent Policy & 0.934 $\pm$ 0.006 & 0.948 $\pm$ 0.002 & 0.940 $\pm$ 0.001 & 0.453 $\pm$ 0.011 & 0.488 $\pm$ 0.002 & 0.474 $\pm$ 0.005 \\
\midrule
\bottomrule
\end{tabular}
\label{tab:agent_ablation}
}
\end{table*}

\begin{figure}[ht]
\centering
\subfigure[Node Classification]{%
    \includegraphics[width=0.48\linewidth]{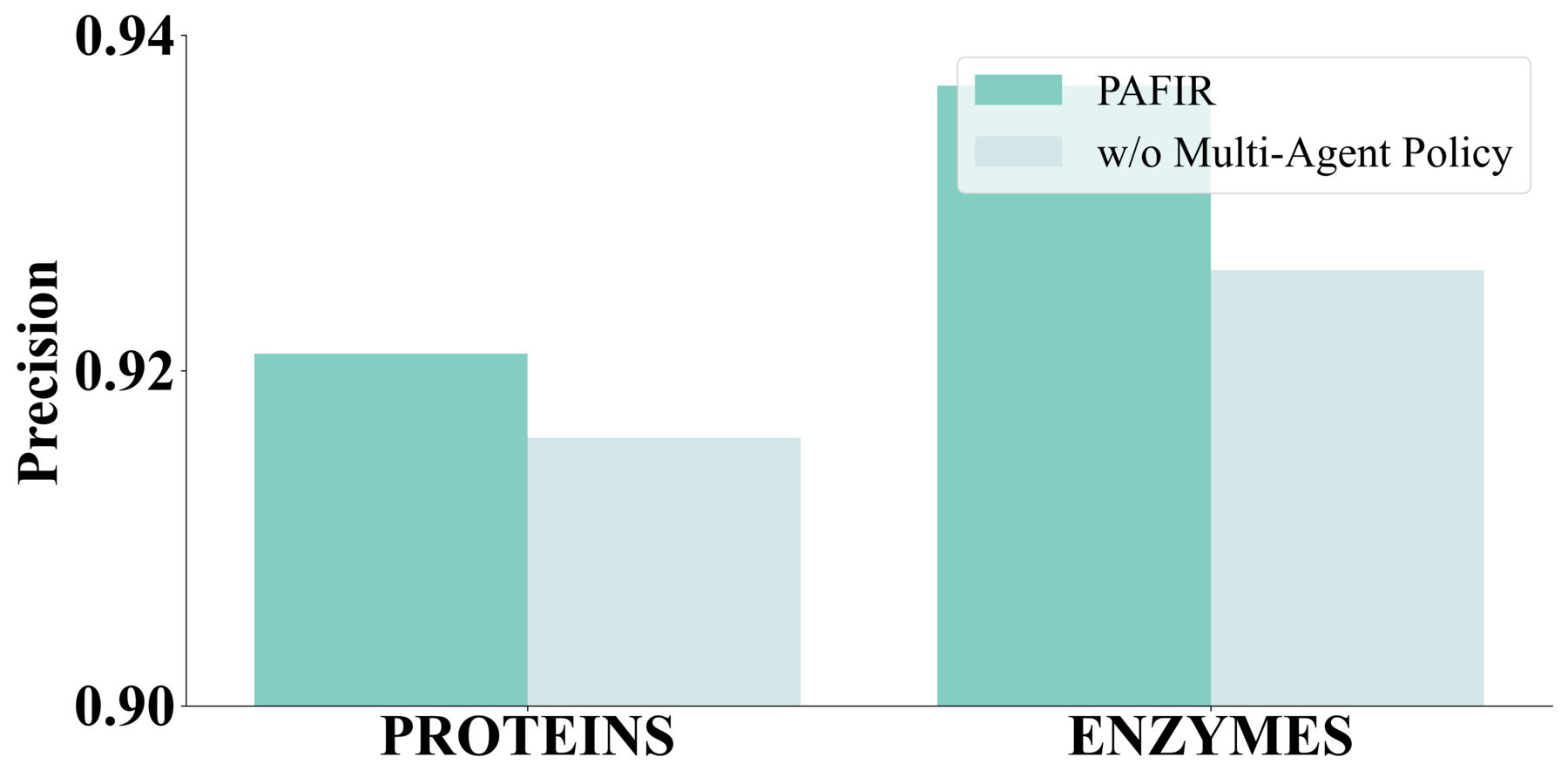}
}
\hfill
\subfigure[Graph Classification]{%
    \includegraphics[width=0.48\linewidth]{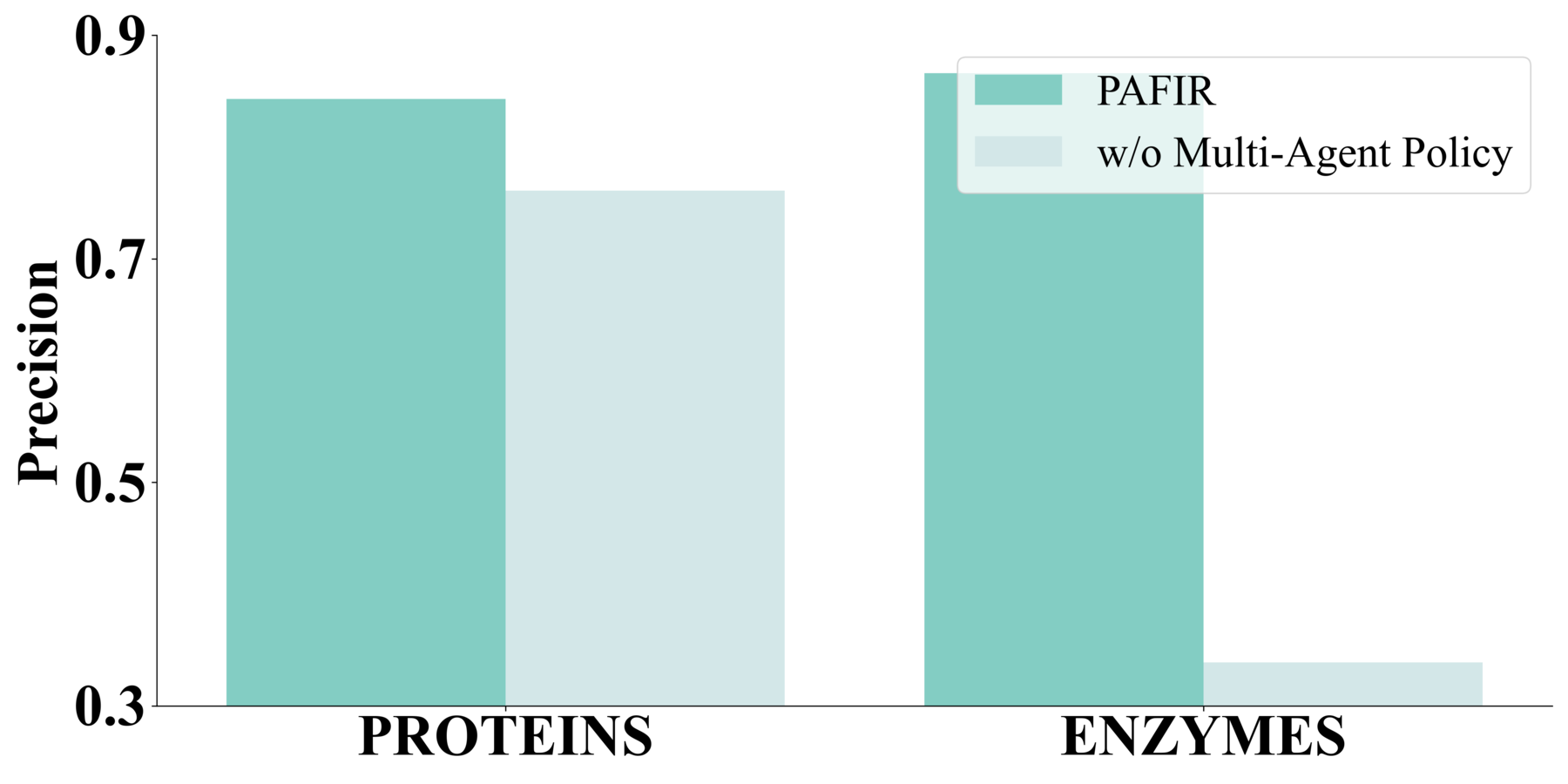}
}
\caption{Ablation experiment on the graph domain. We report Precision under two tasks: node classification and graph classification.}
\vspace{-0.1in}
\label{fig:agent_ablation}
\end{figure}

For example, on the PROTEINS dataset, the F1 score for graph classification drops from 0.832 to 0.759 when the multi-agent policy is excluded. A similar but more severe degradation is observed on the ENZYMES dataset, where the F1 score decreases from 0.857 to 0.306. Similarly, on the AIDS dataset, removing the multi-agent policy leads to a substantial decline in graph classification performance from 0.986 to 0.474. This contrast indicates that the multi-agent strategy plays a critical role in guiding feature transformation and aggregation at the graph level, where complex interactions between feature subsets are more pronounced.

These results confirm that the cooperative mechanism among agents enables PAFIR to better explore and select informative features, ultimately enhancing its capacity for structured representation learning.

\paragraph{Summary of Ablation Findings}
Across all ablation studies, we observe that each core component of PAFIR contributes uniquely and substantially to the model’s overall performance. The hierarchical graph encoder is essential for capturing multi-level structural patterns, with its removal leading to significant performance degradation in both node and graph classification. The time series encoding module plays a key role in modeling temporal patterns; excluding it results in dramatically increased forecasting errors across all benchmarks. Finally, the multi-agent policy proves particularly effective in guiding feature selection at the graph level, as evidenced by the sharp drop in graph classification performance when this component is removed. Collectively, these results demonstrate that the integrated design of structural encoding, temporal modeling, and the multi-agent module in the reinforcement learning framework is key to PAFIR’s potential in learning expressive and generalizable representations across diverse data modalities.

\begin{figure}[ht]
\centering
\subfigure[ENZYMES: Precision]{\includegraphics[width=0.48\linewidth]{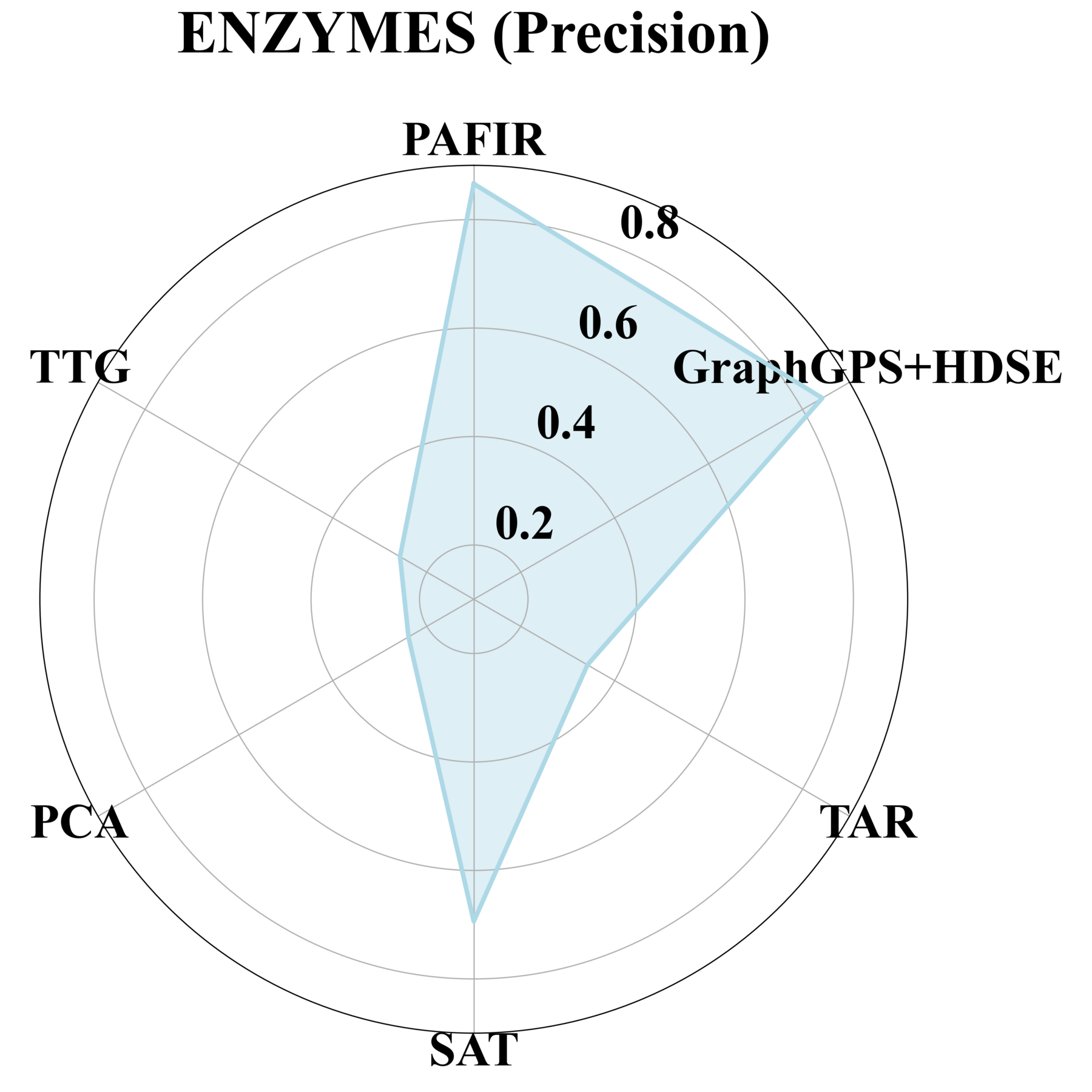} \label{subfig:precision}} 
 \hfill    
\subfigure[TRAFFIC: MSE]{\includegraphics[width=0.48\linewidth]{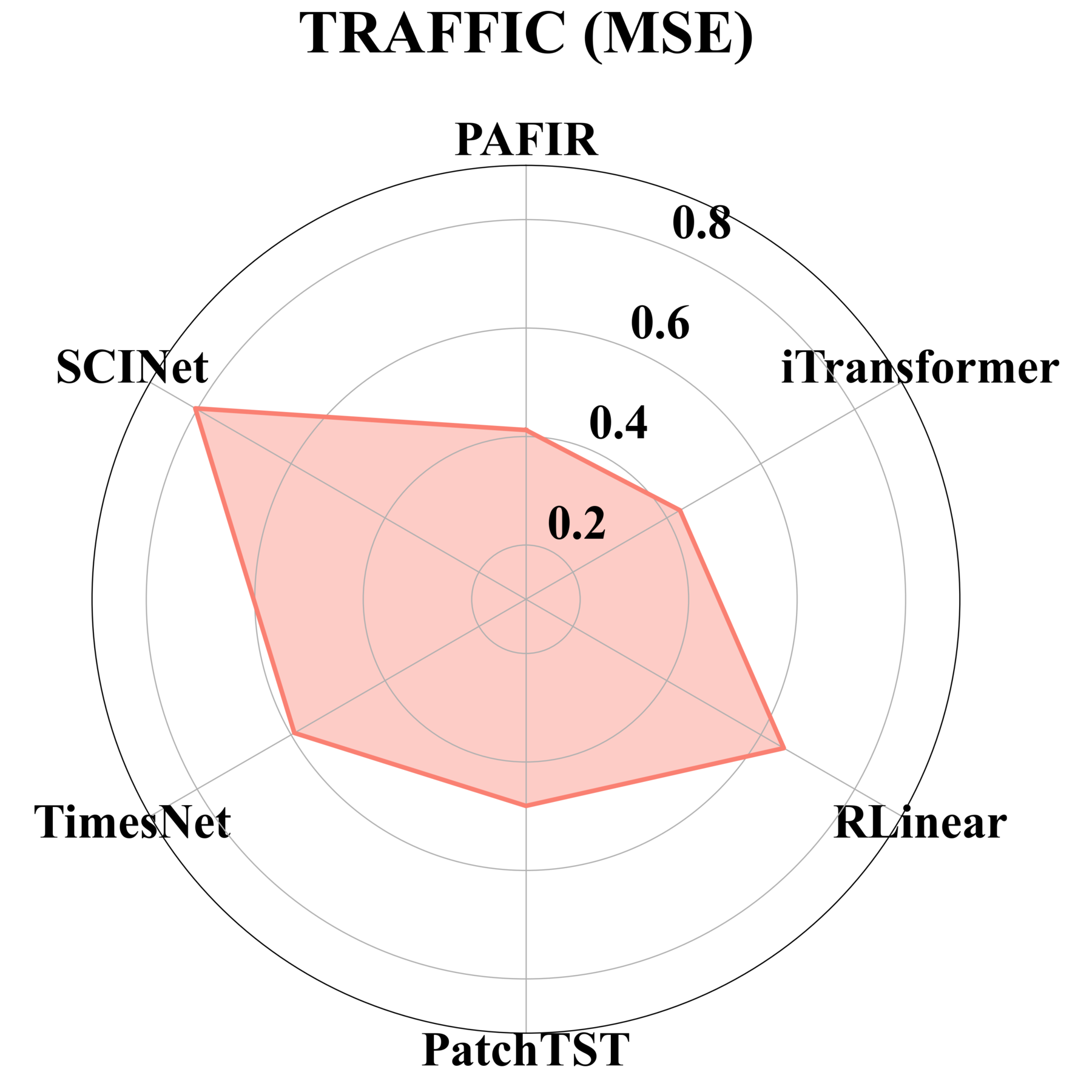}
    \label{subfig:mse}}
\caption{Experiment Result of Different Methods on Two Datasets.}
\label{fig:generalization}
\vspace{-0.3cm}
\end{figure}

\section{Cross-Domain Generality Results} \label{appendix:generality}
We apply the proposed PAFIR framework to two representative tasks, graph-based classification and multivariate time series forecasting, to evaluate its generality across heterogeneous data modalities. Specifically, we test PAFIR on the ENZYMES dataset, which involves structural learning over biological graph data, and the TRAFFIC dataset, which captures temporal dynamics in real-world sensor measurements (Fig.~\ref{fig:generalization}). These benchmarks represent two fundamentally different problem settings and data characteristics, allowing us to assess the robustness and flexibility of our unified framework. These two tasks not only differ in terms of data modality, structured graphs versus sequential time series, but also in the nature of their feature dependencies and temporal or topological patterns. On ENZYMES, PAFIR is compared with state-of-the-art graph learning models, including GraphGPS+HDSE, TAR, SAT, PCA, and TTG. On TRAFFIC, we benchmark PAFIR against competitive forecasting models such as iTransformer, RLinear, PatchTST, TimesNet, and SCINet.

On the ENZYMES dataset (Fig.~\ref{fig:generalization}\subref{subfig:precision}), PAFIR achieves the highest precision among all competing graph-based models. This result highlights PAFIR’s ability to capture discriminative structural features and maintain high predictive accuracy in complex biological graph classification tasks. On the TRAFFIC dataset (Fig.~\ref{fig:generalization}\subref{subfig:mse}), where the evaluation metric is MSE, PAFIR again outperforms temporal forecasting baselines. Its consistently lower MSE reflects the model’s robustness in capturing temporal dependencies and adapting to traffic dynamics.

The consistent gains across both structural and temporal settings highlight the effectiveness of PAFIR’s unified design in handling heterogeneous data. This cross-domain generality underscores its practical value in real-world scenarios where multimodal health, behavioral, or sensor data are commonly encountered.

\end{document}